# COfEE: A Comprehensive Ontology for Event Extraction from text

Ali Balali[a,*], Masoud Asadpour[a], Seyed Hossein Jafari[a]
[a] *School of ECE, College of Engineering, University of Tehran, Tehran, Iran*

## Abstract

Data is published on the web over time in great volumes, but majority of the data is unstructured, making it hard to understand and difficult to interpret. Information Extraction (IE) methods obtain structured information from unstructured data. One of the challenging IE tasks is Event Extraction (EE) which seeks to derive information about specific incidents and their actors from the text. EE is useful in many domains such as building a knowledge base, information retrieval, summarization and online monitoring systems. In the past decades, some event ontologies like ACE, CAMEO and ICEWS were developed to define event forms, actors and dimensions of events observed in the text. These event ontologies still have some shortcomings such as covering only a few topics like political events, having inflexible structure in defining argument roles, lack of analytical dimensions, and insufficient gold-standard data especially in low-resource language such as Persian. To address these concerns, we propose an event ontology, namely COfEE, that incorporates both expert domain knowledge, previous ontologies and a data-driven approach for identifying events from text. COfEE consists of two hierarchy levels (event types and event sub-types) that include new categories relating to environmental issues, cyberspace, criminal activity and natural disasters which need to be monitored instantly. Also, dynamic roles according to each event sub-type are defined to capture various dimensions of events. In a follow-up experiment, the proposed ontology is evaluated on Wikipedia events, and it is shown to be general and comprehensive. Moreover, in order to facilitate the preparation of gold-standard data for event extraction, a language-independent online tool is presented based on COfEE. A gold-standard dataset annotated by 10 human experts is also prepared consisting 24K news articles in Persian language according to the COfEE ontology. In order to diversify the data, news articles from the Wikipedia event portal and 100 most popular Persian news agencies between the years 2008 and 2021 is collected. Finally, we present a supervised method based on deep learning techniques to automatically extract relevant events and corresponding actors.

## 1 Introduction

With the development of the Internet, billions of users publish their opinions and observations about the world on social media and online news agencies. This data provides rapid and geographically distributed information in real-time and has been utilized for monitoring real-world events [1]. Understanding this data can improve the quality of social activities e.g. the international situation awareness, conflict resolution, noticing possible crises, future policy planning, etc. [2]. Nevertheless, most of the aforementioned data is unstructured text, thus hard to analyze [3, 4]. On the other hand, dealing with such a large amount of data is not easy, making it impractical to derive valuable information by manual reading and analysis [2]. By growth of computing power and advances in machine learning, automatic event extraction (EE) has become a hot research topic in natural language processing [2, 5]. It focuses on automatically extracting specific information

[*] This is to indicate the corresponding author.
Email address: balali.a67@ut.ac.ir (A. Balali), asadpour@ut.ac.ir (M. Asadpour), jafari.h@ut.ac.ir (H. Jafari)

about incidents, which appear in texts and discover involved actors, what the event is, when, where and how it has happened.

EE can be useful in variety of domains including question answering [6, 7], summarization [8, 9], news recommendation [10, 11], building a knowledge base automatically based on extracted information [12, 13] which can be a valuable asset for analyzing economic, political and social trends [5], post event rapid damage assessment [1, 14, 15], and online monitoring systems to construct indicators for early warnings in different issues such as natural disaster, health, disease, cyber-attacks, stock markets, accidents, robbery and conflicts [5, 16-24]. Also, the output of EE is a powerful source of information for predictive analyses [5, 25, 26]. Services like RecordedFuture [25], ICEWS [27], EMBERS [28] and GDELT [29] have been utilized to forecast domestic political crises within countries [28]. For example, protest forecasting systems could help with prioritizing citizen grievances [26].

Figure 1 illustrates a subset of important tasks in IE. Named entity recognition (NER) is essential as a preprocessing phase for EE. Moreover, temporal extraction and co-reference resolution can improve the effectiveness of EE approaches.

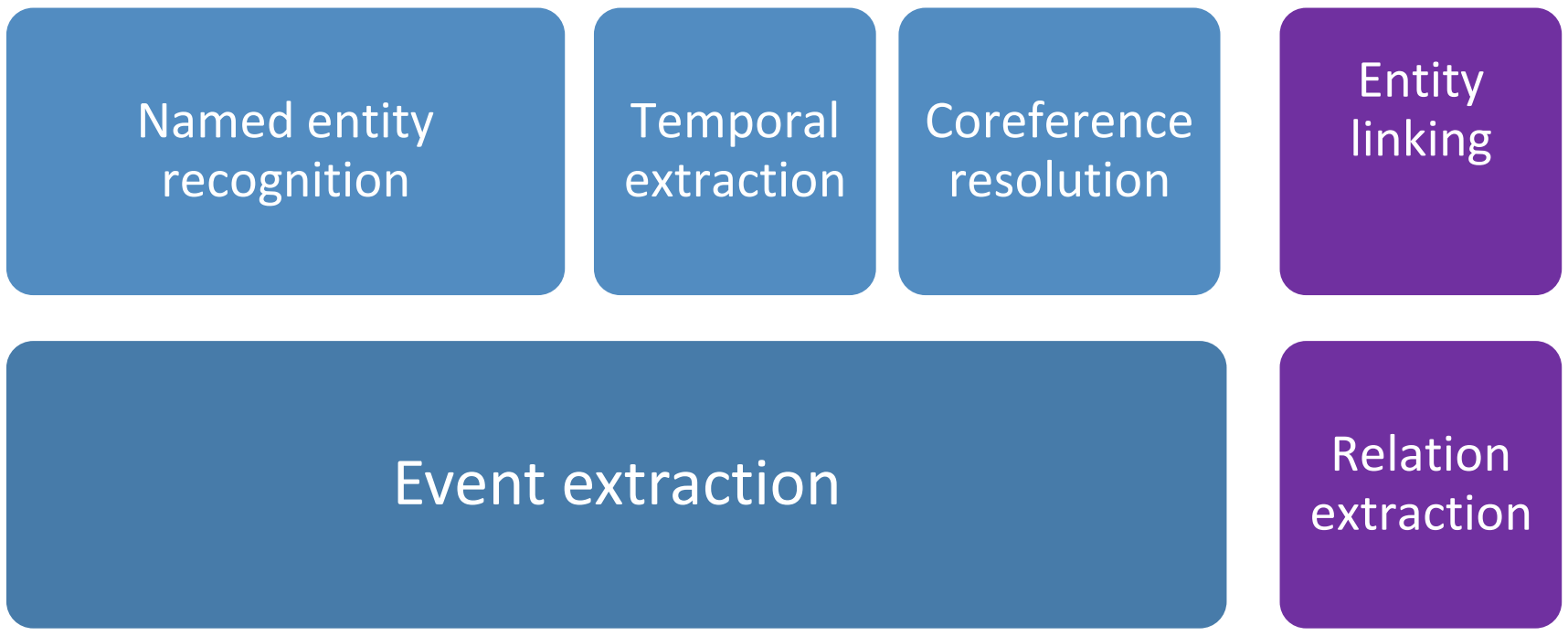

*Figure 1: Some of the top-level information extraction tasks*

In the first step of EE, we need an Event Ontology (EO), which determines what event we are looking for and who their actors are. The first attempts towards EO was conducted in 1960s by DARPA a.k.a. WEIS[1] project [30]. Subsequently, other ontologies emerged, most notably, the COPDAB[2] [31], MID[3] [32], IDEA[4] [21], ACE[5] [33], CAMEO[6] [34], TAC[7] KBP [35], ICEWS[8] [36] and PLOVER[9] [37]. Among these, one of the most popular EOs is ACE, which has been employed to a large extent by many researchers since 2004 [38-41]. In related works section, we discuss these EOs in more details.

---

[1] World Event/Interaction Survey

[2] The Conflict and Peace Data Bank

[3] Militarized Interstate Disputes, http://cow.dss.ucdavis.edu/data-sets/MIDs

[4] Integrated Data for Events Analysis

[5] Automatic Content Extraction

[6] Conflict and Mediation Event Observations

[7] Text Analysis Conference

[8] Integrated Crisis Early Warning System.

[9] Political Language Ontology for Verifiable Event Records

In the following lines, the concept of event trigger, event type and event argument are introduced according to ACE ontology to determine who do what to whom, where, and how in the text. "Event trigger" is a word or a phrase that clearly expresses the event's occurrence. Event triggers can be verbs, nouns, and occasionally adjectives like "dead" or "bankrupt" [42]. "Event type" (what) describes the topic of the event trigger [30]. "Event arguments" are entities that fill specific roles in the corresponding event trigger [42]. The general roles are time (when), location (where), source (who: the initiator of the event), target (whom: the recipient of the event) and instruments (how: with what methods) [30, 31]. Also, there are many other specific roles in event ontologies such as sentence, artifact etc. In Figure 2, an example of EE is illustrated for the sentence ''*A cholera outbreak has since April 27 killed at least 115 people and left another 8,500 ill across Yemen.*'' taken from Wikipedia event portal[10]. In this example, there are three event triggers: "*outbreak*" (rectangular green box), "*killed*" (yellow box) and "*ill*" (blue box) and their corresponding subtypes "*Epidemics*" (which belongs to the environment type), "*Death*" (life type) and "*Injury*" (life type), respectively. Furthermore, there are seven entities "*A cholera*", "*since April 27*", "*115*", "*people*", "*another*", "*8500*" and "*Yemen*" shown in brackets. Arcs in the sentence depict the arguments and corresponding roles associated with each event trigger. For instance, "*since April 27*" and "*8,500*" (see the blue row) are arguments of the event trigger "*ill*" with roles "*Time*" and "*Number of Participants*", respectively.

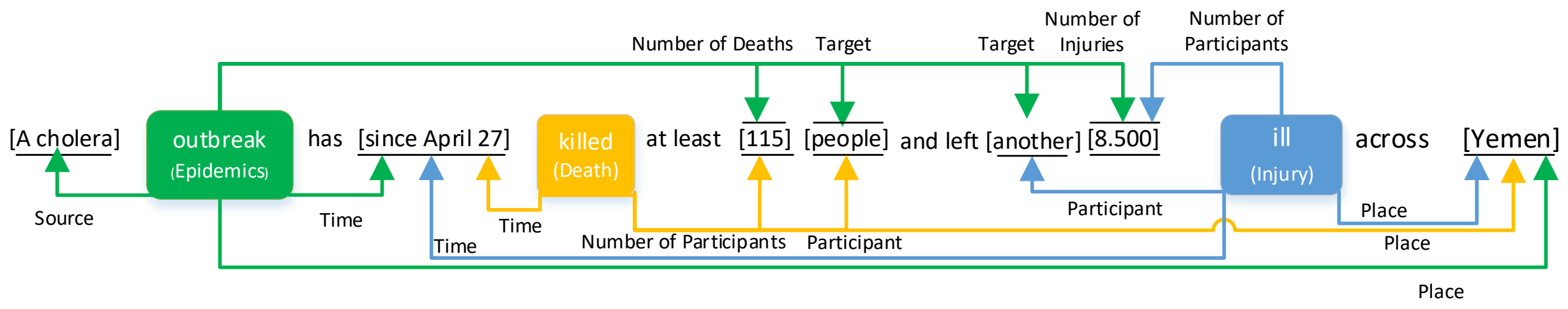

*Figure 2: An example of Event Extraction task*

The output of EE task extracts tuples with the form (event trigger, event type / event subtype, [list of arguments], [list of argument roles]). Event types, event subtypes and roles will be drawn from a specified ontology. In the example on Figure 2 we use our ontology which is one of the main contributions of this paper.

Despite considerable advances in recent years, there are several challenges in the existing event ontologies, most notably:

- **The complexity in choosing the type of events,** for example in CAMEO, 204 event subtypes are defined on political issues in a hierarchy of 4 event type levels which makes it impractical for human coders to find the correct type of events in a given sentence. In many cases, distinguishing between subcategories is difficult. This leads to difficulty in expanding the gold-standard data.

[10] https://en.wikipedia.org/wiki/Portal:Current_events

- **Inflexible structure and lack of analytic dimensions** [21], for instance, CAMEO and ICEWS have defined fixed argument roles including "*Source*", "*Target*", "*Place*" and "*Time*" representing all event subtypes. Lack of different roles for event subtypes leads to loss of information (specifically event dimensions) such as "*number of deaths"*, "*number of injuries"* etc.
- **Most event ontologies only cover political event types** [43] such as WEIS, CAMEO and ICEWS. Thus, we **need new event types**. In other words, some of the important event types are not addressed such as environmental issues, accidents and cyberspace attacks in these event ontologies. In ACE, definition of the events are **limited and coarse-grained** in a sense, therefore making it less suitable for practical applications [44].
- **Lack of suitable tools to create gold-standard data** according to event ontologies. A tool is needed that is multilingual and easily extensible to new domains [5].
- **Scarcity of sufficient gold-standard data** in different languages especially in low-resource language such as Persian. ACE and TAC, which are the most popular datasets in EE, contain only 599 and 360 documents in English respectively, which too few for machine learning approaches that employ numerous learning parameters [45-47].

To address these concerns, we propose a new event-ontology, namely "*COfEE*" (which stands for a Comprehensive Ontology for Event Extraction), with a more general event schema suitable for practical applications that is simpler, more flexible and it is inspired by previous event-ontologies, such as ACE. Moreover, for more clarity, each event subtype is accompanied by examples and descriptions.

In COfEE, 12 event types and 119 event subtypes in two hierarchy levels are introduced. The event subtypes are defined such that they are conceptually different from each other (with the least possible overlap) in order to solve the **complexity in choosing the type of events.** COfEE covers a wide range of topics, from "Politics" to "*Life*", "*Business*" and "*Natural Disasters*". It also **defines new event types** such as "*Cyberspace*", "*Science*" and "*Environment*", and new subtypes such as "*Cyberattack*", "*Extinction*" and "*Epidemics*" by incorporating a data-driven approach and domain knowledge of human experts. We assess the generalization of the proposed event subtypes on a dataset of Wikipedia events of the past 10 years, gathered by multiple individuals with different perspectives throughout the years. According to each event subtype, dynamic roles are defined in COfEE to capture information about dimensions of an event such as "*Number of deaths*", "*Number of injuries*" as a means to tackle the **inflexible structure and lack of analytical dimensions problem.** In order to prepare gold-standard data, we present **a web-based tool** with user-friendly interface to speed-up the process of annotating data and reducing its complexity. This tool is multilingual and easily extensible to new domains. Then, we prepare the first gold-standard data with about 24K sentences for EE in Persian language according to the COfEE event ontology. In order to have a diverse and comprehensive dataset, we have collected news articles from various sources such as the Persian Wikipedia event portal[11] and 100 most popular Persian online news agencies. We used headlines rather than the entire news text in the annotation phase because the headline is short, rich and representative of the main body of the article. To annotate the collected news articles, 10 human experts utilized the COfEE annotation tool.

Our key contributions of this paper lie in the following six aspects:

- We develop a new event ontology which covers many contemporary issues with a flexible structure with defined roles for each event subtype. COfEE covers 12 event types and 119 event subtypes in two hierarchy levels with 21 argument roles. This event ontology introduces new event types such as "*environment*", "*science*", "*cyber-space*" and subtypes such as "*smuggling*", "*economic corruption*", "*kidnapping*", "*extinction*", "*epidemics*" etc. Additionally, COfEE presents new roles such as "*scale*", "*number of missing entities*" and "*number of participants*" to better capture different dimensions of events.
- In definition of event subtypes, we have incorporated a data-driven approach along with the study of previous ontologies for fabricating a more practical and general-purpose event ontology.
- We study the generalization of our proposed framework on a corpus of Wikipedia events and show that it can cover a large proportion of its events.
- We prepare an open-source multi-user web-based language-independent tool for manual event extraction according to the proposed event ontology which can be easily extended to new domains. The online version of this tool is available on cofee.sociallab.ir and the source code is published on GitHub[12].
- We prepare the first gold-standard data for EE with about 24K sentences from different sources in Persian, annotated by 10 human experts according to COfEE event ontology. This dataset is available on GitHub.
- We propose a supervised method to automatically extract events based on the mentioned dataset as a baseline approach.

The remainder of this paper is structured as follows: Section 2 describes the related works, where we review some popular event ontologies and compare them with each other. In the next section, the proposed event ontology is introduced. We provide a collection of entities, event types and subtypes, as well as argument roles. In Section 4, COfEE tool is presented for manual coding (annotation) of events in text. Section 5 provides statistics on the prepared dataset. Section 6 describes our deep learning-based method to automatically extract events and their corresponding actors. The generalization of COfEE ontology and the experiments results of the proposed method on the prepared dataset are discussed in section 7. Finally, Section 8 concludes the paper.

[11] https://en.wikipedia.org/wiki/Portal:Current_events
[12] https://github.com/utsnlab/COfEE

## 2 Related Works

During the twentieth century, two event ontologies COPDAB[13] [31] developed by Edward Azar and Charles McClelland's WEIS [48] dominated event data research. These two EOs have focused on international politics, especially official diplomacy and military threats. Human coders were meant to read news stories and code events based on these EOs, that was very costly [49]. For four decades, these EOs were used with only minor modifications. However, these EOs could not capture contemporary issues such as ethnic disputes, organized crimes and violence. During the early 2000s, IDEA [21] and CAMEO [50] were introduced as new political EOs. The IDEA framework is a refinement and an extension of WEIS. IDEA provides a comprehensive framework for events which is useful to monitor contemporary trends in civil and inter-state politics [21]. Also, CAMEO has served as the basis for most of the modern event datasets such as the Integrated Crisis Early Warning System (ICEWS). By the mid-2000s with a fund from DARPA, ICEWS[14] the next generation EO emerged with the goal of automated event extraction instead of manual event extraction. ICEWS corrected some of the recognized ambiguities in WEIS and COPDAB. ICEWS event types were originally derived from CAMEO whereas it also captured the events related to domestic political crises within countries [28, 51].

In 2004, a research program called ACE[15] was conducted for developing advanced information extraction technologies. In this program, an event ontology along with the corresponding dataset was proposed [42]. ACE has largely been used by researchers to automatically construct models for event extraction from the text. In 2009, ACE program participated in Text Analysis Conference (TAC) that was focused on particular sub-problems such as "Event Nugget Detection". The goal was to extract information that is suitable as input to a knowledge base. ACE and TAC are currently the most popular and the most cited datasets for the EE task. In 2018, a simplified version of CAMEO was introduced under the name PLOVER. PLOVER added a "mode" field for each event type which summarized 204 subtypes of CAMEO and made it easier to work with.

Table 1 compares some of the most popular EOs according to their year of development, covered issues, number of event types and subtypes, number of argument roles and the amount of gold-standard data. According to this table, all EOs except IDEA, ACE and TAC cover only the political issues, indicating the need for **new event types** for covering other issues. The number of event subtypes in IDEA, CAMEO and ICEWS are more than 200, presented in four hierarchy levels. Appearance of too many event subtypes focusing on only political issues leads to **complexity when choosing among them** [37]**.** The variety in argument roles can help in capturing different dimensions of events. Except ACE, TAC and Plover, other EOs have a pre-defined set of roles for all event type and subtypes causing **inflexible structure and lack of analytical dimensions.** For example, the output of CAMEO and ICEWS consists of three component parts: (*Source Actor*, *Event Type*, *Target Actor*) as well as general attributes e.g., *time* and *location*. These formats cannot provide some of the useful information available in texts such as "*number of participants*", "*number of deaths*" etc. that are particularly useful for early warnings and assessing the dimensions of events.

[13] The Conflict and Peace Data Bank
[14] Integrated Conflict Early Warning System
[15] Automatic Content Extraction

*Table 1. Comparison of the most popular event ontologies*

| Event Ontology | Year | Covered Topics | # Event Types and Subtypes | Argument Roles | Size of Gold-standard Data |
|---|---|---|---|---|---|
| **WEIS [30]** | 1967 | Official diplomacy and military threats | 22 events and 63 subevents | 4 argument roles: Actor, Target, Time and Location. Fixed roles for all event subtypes | NA[16] |
| **COPDAB [31]** | 1975 | Political and international interaction | 16 events | 7 argument roles: Actor, Target, Source, Time, Scale value, the Verbal and Physical Act, Location. Fixed roles for all event subtypes | NA |
| **IDEA [21]** | 1998 | Social, economic and political events data | 38 events and 245 subevents | 4 arguments roles: Source and Target, Time and Location. Fixed roles for all event subtypes | NA |
| **MID [32]** | 1996 | Displays of force, threats of force and uses of force | 3 events and 20 subevents | 21 arguments roles: Hostility level, Fatality level etc. Fixed roles for all event subtypes | ~2436 Records [17] |
| **CAMEO [34]** | 2000 | Political events | 20 events and 204 subevents | 4 arguments roles: Source and Target, Time and Location. Fixed roles for all event subtypes | NA |
| **ACE [42]** | 2004 | Life, movement, transaction, business, conflicts, contact, justice and personnel. | 8 events and 33 subevents | 35 arguments roles: Agent, Attacker, Artifact, Seller, Buyer etc. Dynamic roles for each event subtype | ~20000 sentences in 599 documents |
| **TAC [35]** | 2015 | Life, movement, transaction, business, conflicts, contact, justice, manufacture and personnel. | 9 events and 38 event subtypes | Similar to ACE. | ~14000 sentences in 360 documents |
| **ICEWS [36]** | 2007 | Political events (expanded version of the CAMEO) | 20 events and 312 subevents | 4 arguments roles: Source and Target, Time and Location. Fixed roles for all event subtypes | NA |
| **PLOVER [37]** | 2018 | Political events (expanded version of the CAMEO) | 18 event types with mode fields | 4 arguments roles: Source and Target, Time and Location Fixed roles for each event subtype | 732 Sentences[18] |

[16] Not available
[17] Http://cow.dss.ucdavis.edu/data-sets/MIDs
[18] Https://github.com/openeventdata/PLOVER/

The amount of gold-standard data is an important feature in EOs. Unfortunately, according to Table 1, only a few EOs have published gold-standard data. To automatically extract events, **scarcity of gold-standard data** is an obstacle in using machine learning approaches, especially in deep learning. In practice, only the event extraction data provided by ACE (~20K sentences in 599 documents for English) [42] and TAC (~14000 sentences in 360 documents for English) [52] for English, Chinese and Arabic are sufficient for basic machine learning methods [45, 46, 53-56]. Deep learning techniques with numerous parameters, however, will likely require more data than this [45-47]. Furthermore, many of the event subtypes in ACE and TAC's documents have similar events (containing little new information) and also, less than 30% of sentences in ACE consist of an event. Some event subtypes such as “extradition” and “pardon” appeared in few sentences. For languages with limited resources, such as Persian, Portuguese, etc., there are no event extraction datasets available. As a result, we need to prepare gold-standard data to develop language-specific event extraction tasks. In order to achieve this goal, tools are required for preparing gold-standard data according to EOs. A suitable tool can increase the speed and accuracy of preparing the gold-standard data. Unfortunately, these EOs do not provide official annotation tools customized for the ontology. The process of event extraction annotation has several phases that can be tremendously facilitated by leveraging a well-designed capable tool. Instead, approaches based on designing various handcrafted patterns have grown large. Designing these patterns takes a large amount of human effort while suffering from generalization deficiency. Two popular software in this context are: 1) TABARI (Textual Analysis By Augmented Replacement Instructions)[19] according to WEIS that uses sparse parsing to recognize patterns in text, and 2) Petrarch (Python Engine for Text Resolution And Related Coding Hierarchy)[20] based on CAMEO which extracts events and actors by fully-parsed Penn TreeBank to deal with syntactic and grammatical structures [49]. These software extract arguments by looking up in the verbs dictionary, the actors dictionary and the issues dictionary. Since these software need a set of language-dependent tools, such as treebanks and dictionaries, it is difficult to develop them in other languages.

As we need to cover events which belong to a variety of topics as well as extracting different dimensions of events, the proposed EO is more inspired by ACE and TAC. In the proposed event ontology, we extend ACE and TAC event types and subtypes to cover new important contemporary topics and analytical dimensions, in addition to correcting some of their existing ambiguities. In the next section, the proposed EO is introduced in more details.

## 3: COfEE event ontology

In the current section, we introduce an ontology for event extraction, “*COfEE*”, that includes a list of entities, event types and argument roles. Also, in order to better understand the EO, diverse examples for each event subtype is provided.

[19] Http://eventdata.parusanalytics.com/software.dir/tabari.html
[20] Https://petrarch2.readthedocs.io/en/latest/petrarch2.html

## 3-1: COfEE Entities

Entity (usually a noun phrase (NP), e.g., "*Barak Obama*, "*Yemen*", "today") is an object or a set of objects in the world. The entity attributes include the name(s) used to refer to the entity and the entity type which refers to the class of the entity. The possible entity types in COfEE are listed in Table 2.

In case there are multiple entities in the same phrase, we always use a rule of thumb that considers the most general entity. For example, the phrase *"President Barack Obama"* can be broken down into two entities – *"Occupation"* (President) and *"Person"* (Barack Obama) – we take the entire phrase as one entity of the type "*Person*".

*Table 2. COfEE entity types*

| Entity Types |
|---|
| **Organization:** e.g., Iran's Guards, West Midlands Police, The United States Senate |
| **Geo-Political Entity:** Nation, Continent, County-or-District, State-or-Province |
| **Location:** Address, Boundary, Region |
| **Person:** e.g., Putin, Palestinian President Mahmoud Abbas, President Donald Trump |
| **Animal:** e.g., The South China Tiger, The Javan Rhino, Elephants |
| **Facility (objects, man-made structures):** e.g., The bomb, A car, Oil, Diesel fuel, Software, Hospital, Gym |
| **Time:** e.g., Sunday, The 1970s, Last year |
| **Numeric:** e.g., 100, 20$, One thousand |
| **Occupation:** e.g., A student, Architect, Accounting manager, President |
| **Contact-Info:** URL, Email, Phone Number |
| **Disease:** e.g., Cholera, Corona, Covid-19, Polio |

There are many NER tools [57, 58] in different languages with an accuracy of more than 80%, which could be useful to detect entities in text. However, if there is no proper tool in specific languages, the following options can be employed to solve the problem:

- Preparing dictionaries for each entity type and then looking a phrase up. Also, using a rule-base tool for extracting number, time and contact-info entities. The title of the Wikipedia pages can help in building these dictionaries.
- Using COfEE tool to annotate entities and entity types manually by human annotators.

## 3-2: COfEE Event Types and Subtypes

One of the event extraction steps is to identify event triggers and to classify them into their corresponding type (e.g., "*Crime*") and subtype (e.g., "*Attack*"). Event triggers serve to answer questions such as "*what happened*" by means of a verb (e.g., "*kill''*), a noun (e.g., "*earthquake*"), and, occasionally through adjectives like "*loser*" (e.g., the loser party). In COfEE, event trigger attributes include the event *type*, event *subtype*, *tense*, *modality* and *polarity*. COfEE introduces 12 event types and 119 subtypes in different issues listed in Table 3. The tense, modality and polarity properties will be covered in Section 3-4.

We have incorporated a data-driven approach along with the study of previous ontologies for defining event subtypes. In one phase, some of the most important event ontologies were scrutinized and the event types that met our conditions (those of greater importance e.g., events that require constant online monitoring) were handpicked. If a subtype could be assigned to multiple types, the most common use case in the domain was considered. *"Meeting"*, for example, can both be found in *"Politics"* and "*Business*", but it appears more frequently under "*Politics*" events in the news. In the next step, the most frequent n-grams (1-grams to 5-grams) of randomly selected news headlines were extracted and reviewed by human experts. Finally, these n-grams were selected out by the experts to extend the originally hand-picked events, resulting in a more general event ontology that can be used by other researchers and practitioners.

*Table 3. COfEE event types and subtypes*

| Event type | Event subtypes |
|---|---|
| **1) Life** | Death, Injury, Birth, Drowning, Survival, Marriage, Divorce, Hospitalization, Missing, Immigration, Suicide |
| **2) Natural Disasters** | Volcanic Eruption, Tsunami, Earthquake, Landslide, Avalanche, Bad Weather, Storm, Flood, Drought |
| **3) Environment** | Pollution, Hunting, Extinction, Epidemics, Emergency Evacuation, Resource Shortage, Quarantine |
| **4) Crime** | Attack, Explosion, Hostage Crisis, Sex Assault, Kidnapping, Homicide, Torture, Escape, Smuggling, Robbery, Economic Corruption, Destruction, Espionage, Copyright Violation, Human-Rights Violation, Privacy Violation |
| **5) Justice** | Complaint, Arrest, Pardoning, Prosecution, Execution, Imprisonment, Fining, Trial, Surrender, Prisoner Release |
| **6) Business** | Start Position, Recruitment, End Position, Money Transfer, Release, Ownership Transfer, Pricing, Establishment, Bankruptcy, Produce, Investment, Price Rise, Price Drop, Production Rise, Production Drop, Initial Public Offering , Capital Increase |
| **7) Politics** | Travel, Aid, Cooperation, End Cooperation, Meeting, War, End War, Conflict, Ban, Sanction, Threat, Conquering, Occupy, Extradite, Exile, Apologize, Deport, Interference, Impeachment, Export, Import, Dissolution, Condemnation, Troops Withdrawal, Suppress, Military Exercise, Criticism, Refuge, Settlement, Breach of Settlement, End Settlement |
| **8) Social** | Protest, Coup, Ceremony, Revolution |
| **9) Cyberspace** | Cyber Attack, Information Disclosure |
| **10) Election** | Election Candidacy, Election Results, Holding Election |
| **11) Accident** | Rail Accidents, Marine Accidents, Aviation Accidents, Traffic Collision, Collapse, Hazardous material spill, Fire |
| **12) Science** | Discovery, Invention |

## 3-3: COfEE Event Argument Roles

Arguments are the entities which have specific roles in events. There are 21 different roles defined to be assigned to entities in COfEE, as listed in Table 4. Also, argument roles for each event subtype along with examples are listed in Table 8. For each event subtype, the corresponding argument roles are presented. For example, the event subtypes in the event type "*Natural*

*Disasters*” has a role “*scale*” which does not show up in other event types. In Table 8, event subtypes are grouped according to their roles. It should be noted that the “*source*” role specifies the agent that originates, causes or initiates the event (the main responsible entity) such as the *Person* entities in event subtype “*Suicide*”. Alternatively, the “*participant*” role specifies an entity that is involved in the event (probably affected by the event) such as the *Person* entities in event type “*Natural Disasters*”. In COfEE, a role can be assigned to multiple entities in the event, but each entity has at most one role in the event.

*Table 4. Argument roles of COfEE*

| Description of Argument Roles | |
|---|---|
| 1) Participant: Who/what is involved or takes part in the event? (not as source or target) | 12) Duration: How long has the event lasted or continued? |
| 2) Source: Who/what is the initiator of the event? | 13) Number of Participants: How many participants are involved? |
| 3) Target: Who/what is the recipient of the event? | 14) Number of Injuries: How many entities (persons, animals) are injured? |
| 4) Time: When does the event take place? | 15) Number of Deaths: How many entities (persons, animals) are dead? |
| 5) Place: Where does the event take place? | 16) Number of Missing Entities: How many entities (persons, animals, Facilities) are missing? |
| 6) Instrument: With what tools does the event take place? | 17) Number of Destructions: How many entities (Facilities) are destroyed? |
| 7) Vehicle: What type of vehicle is used in the event? | 18) Number of Sources: How many sources are affected by the event? |
| 8) Artifact: What is the product of the event (produced, transferred, dealt …)? | 19) Number of Targets: How many targets are affected by the event? |
| 9) Occupation: What is the position of the entity (participant, source or target)? | 20) Scale (Magnitude): What is the scale of the event? |
| 10) Fluctuation: How much has the price of the artifact(s) changed? | 21) Price: How much does the artifact cost? |
| 11) Contact-info: What is the contact information of the entity (participant, source or target)? | |

## 3-4: Event Properties

Events include a few properties giving information on e.g. when and whether the event actually occurred. Currently a simplified version of ACE [59] is used; Polarity, Tense, and Modality properties are tagged. Following, a brief explanation about the values of each property is given:

### <u>Polarity</u>

**Negative** events are those that are explicitly mentioned as not happening. In other words, it is intentionally made clear that they have not happened. Every other event is **Positive**. Two ways of expressing negative polarity are possible: (1) explicitly with negative words, such as “not” or “never”; and (2) implicitly in negative contexts such as denial, refusal or disobedience.

Some negative examples are:

- Trump reiterates his position that Russia did *not* interfere in the 2016 US presidential election.
- Officials in Florida *refuse* to allow the police chief to resign.

**Tense**

Time is determined by comparing the time of publication or broadcast with the time the event took place. It could have four values: **Past**, **Present**, **Future** and **Unspecified** i.e., events that occur after the publication time would be determined as Future.

**Modality**

If an event is **Asserted**, it would be referred to as it really occurred e.g.:

- She traveled to Colorado and purchased a shotgun and ammunition upon arriving.
- A bomb exploded at a downtown bus station in Kenya's capital as passengers boarded a bus.

Other modalities such as *believed, hypothetical, commanded, requested* events etc. will be marked as **Other**. Examples include, but are not limited to:

- *Rumors* of arrests circulated in Vancouver.
- *Should* he not pay the money, they would kill him.
- He was *ordered* to return to Moscow.
- They *wanted* to acquire the company last year.

# 4: COfEE annotation tool

In this section, we introduce the user interface and the annotation process for creating gold-standard data according to COfEE tool. The online free version of this tool is available on cofee.sociallab.ir and the codes can be accessed on GitHub[21].

## 4-1: The annotation process of COfEE tool

In Figure 3, the entire annotation process in COfEE tool is illustrated. In the first step, a user is registered as the admin user. Then, the admin user creates sub-users responsible for text annotation. In the second step, the admin user defines projects. Sub-users can be assigned to different projects and each project can include texts from different domains (topics). In the third step, entity types, event types, event subtypes and argument roles are defined according to COfEE, although these can be further customized by the admin user. In other words, the COfEE tool supports defining new entity types, event type, event subtypes and argument roles. In the next step, the admin user imports texts using web forms, Excel or CSV files into COfEE. It is also possible to import entities that are extracted automatically by NER tools. To facilitate the annotation phase, we suggest the usage of news headlines instead of using the news body, as the headline is shorter and rich enough to represent the main body.

In the fifth step, annotation phase is up and running. For demonstration, a user selects an arbitrary entity, afterwards he/she should select an event trigger that appears on the text and assign the most relevant event subtype to it. Then, an entity is selected and a role is assigned to it according to the corresponding event subtype. In the final step, the admin user can export the data annotated by sub-users in Excel or CSV format.

[21] https://github.com/utsnlab/COfEE

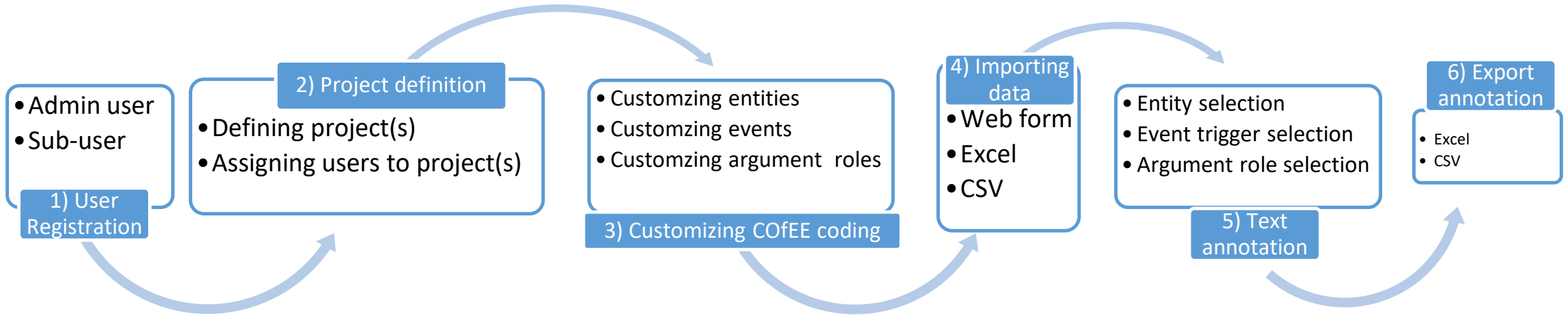

*Figure 3. The entire annotation process of COfEE tool*

## 4-2: The interface of COfEE tool

In to Figure 6, the user interface of COfEE tool in annotation phase is shown.

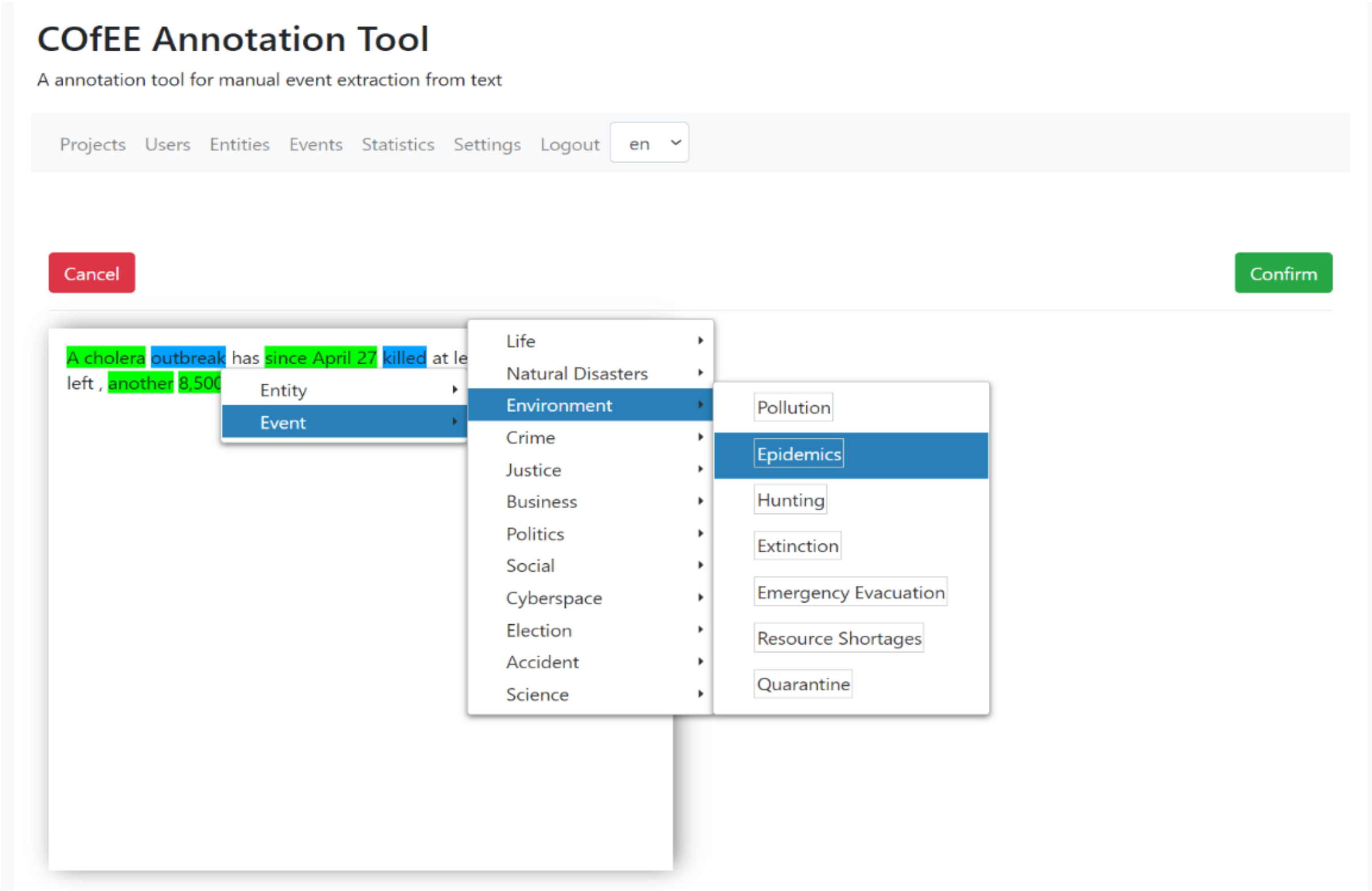

*Figure 4. User interface of COfEE tool in event trigger annotation when "outbreak" is selected as an event of the subtype "Epidemics".*

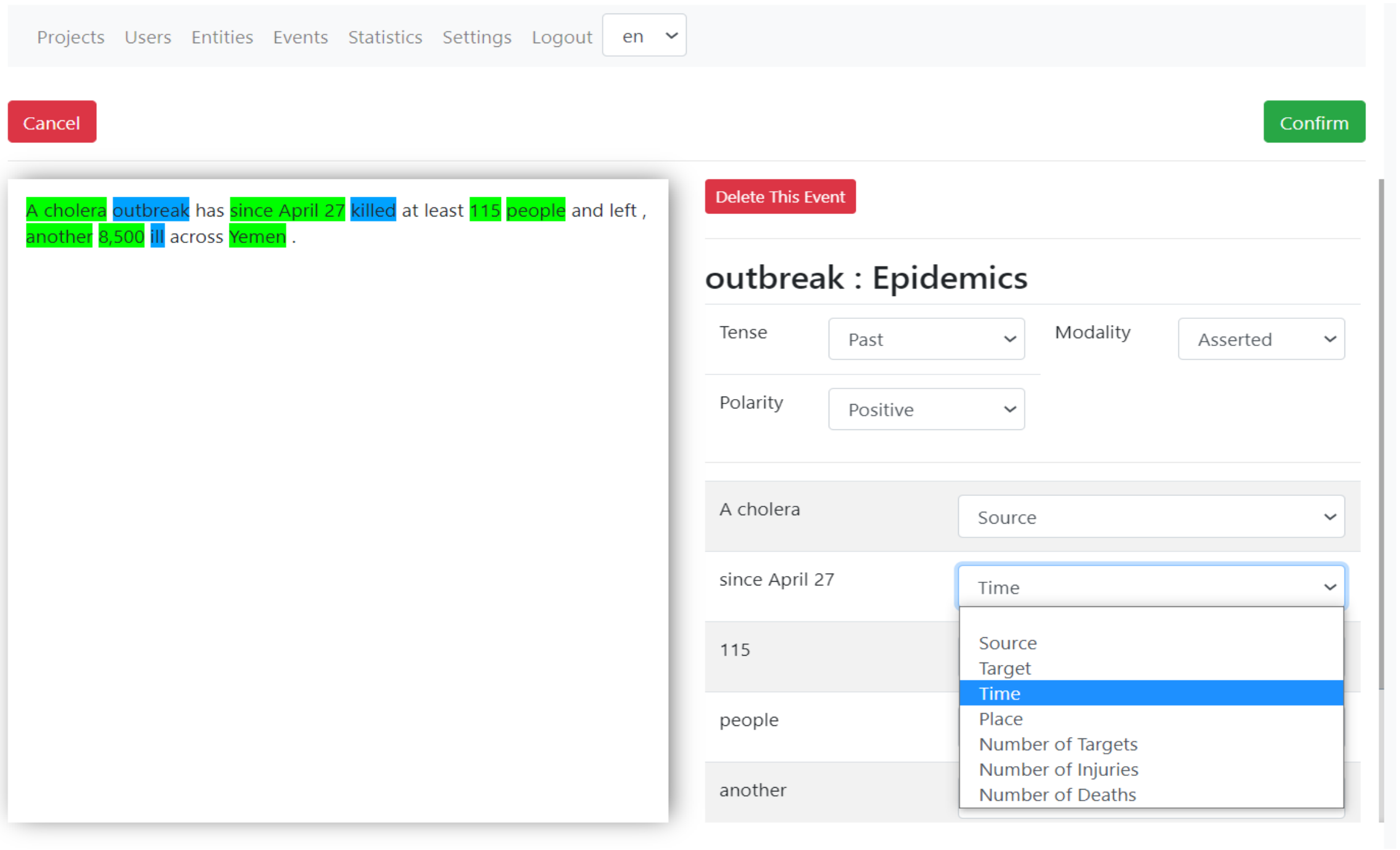

*Figure 5. User interface of COfEE tool in argument roles annotation phase: selecting the arguments of an "outbreak" event. The roles of the entities are selected using a drop-down list.*

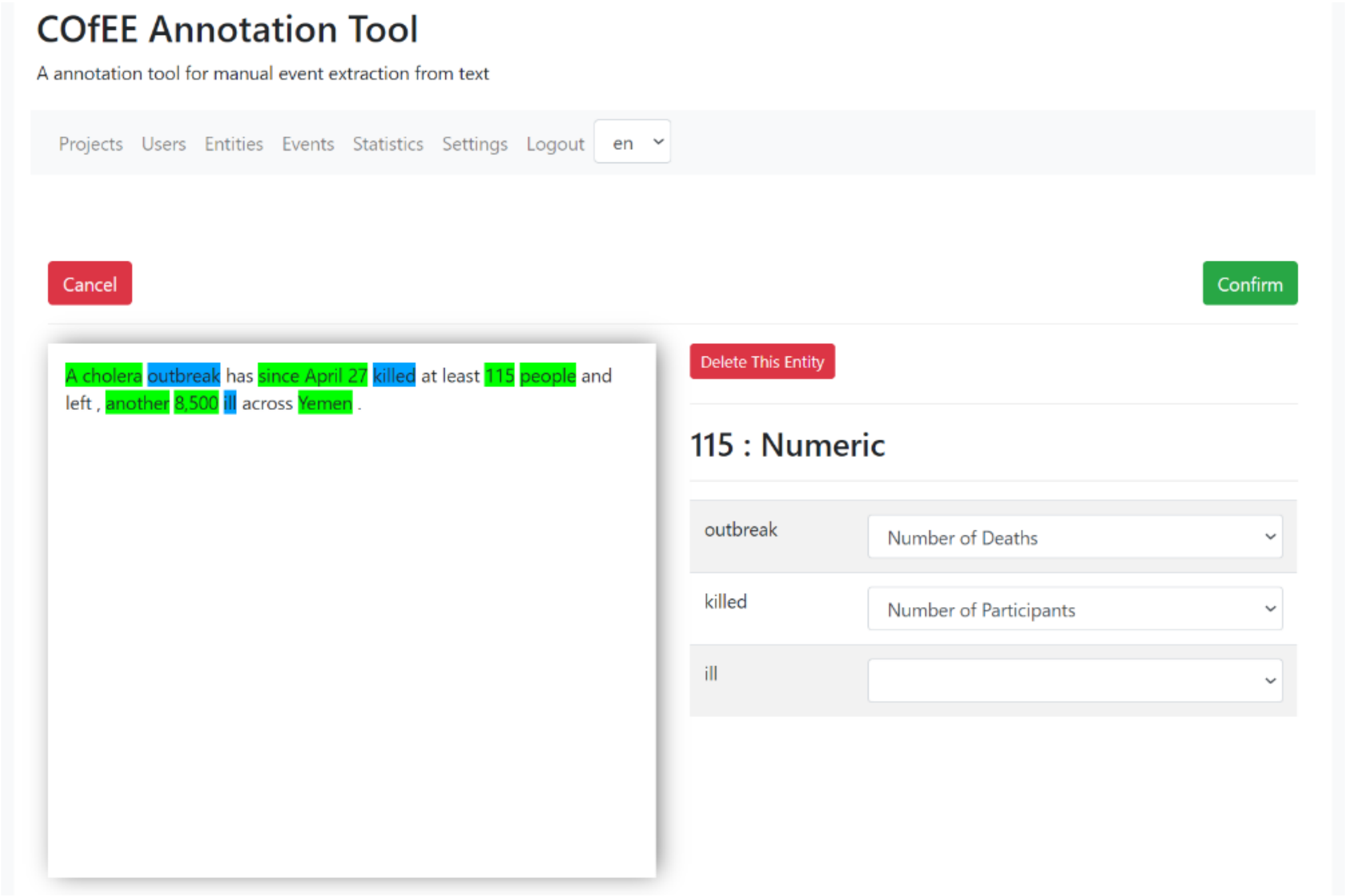

*Figure 6. User interface of COfEE tool where an entity "115" is selected and its roles in different events are depicted.*

The COfEE tool is developed using PHP and the relational database MySQL is used for storing and managing the data. The key features of COfEE tool are:

- COfEE tool supports both RTL languages such as Persian and Arabic and LTR languages such as English and French.
- Users can annotate entities, event triggers and argument roles that appear on the text.
- User interface of COfEE tool is simple and user-friendly, hence it can speed up the annotation step and reduce costs.
- Users can define multiple projects in various domains and assign different users to them.
- By default, entities, event types and argument roles are defined according to COfEE, although, the users can delete, edit them or add new entities, event types and argument roles.
- COfEE is a web-based tool which supports multiple users, handles concurrency issues and incorporates modern capabilities of web applications.

## 5: Dataset Statistics

In order to diversify the news categories, we collect news from different sources including: 1) Persian Wikipedia event portal that lists the most important news stories (containing events) for 13 years between 2008 and 2021. 2) A random sample of the news articles published online during the year 2019 by one of the 100 most popular Persian news agencies and websites (duplicate news were removed) and 3) News articles from these news agencies and websites that contain at least one of the words presented in Table 9 (these words are meant to represent events, we translate them to Persian).

Preparing the dataset for event extraction has several steps (entity recognition, event trigger classification, classification of the role of entity corresponding to event triggers) and therefore it is time-consuming. In order to accelerate the annotation process, we use news headlines which are shorter and richer compared to the main body. The gathered news are manually annotated by 10 human experts using the COfEE annotation tool. Table 5 shows some statistics on the prepared gold-standard data.

*Table 5. The gold-standard data statistics*

<table>
<tr><th colspan="2">Data</th><th>Count</th></tr>
<tr><td colspan="2"># Sentences</td><td>24119</td></tr>
<tr><td colspan="2"># Words</td><td>326455</td></tr>
<tr><td colspan="2"># Entity Mentions</td><td>65612</td></tr>
<tr><td colspan="2"># Event Triggers</td><td>28393</td></tr>
<tr><td rowspan="4"># Event Triggers (by Tense)</td><td>Past</td><td>18372</td></tr>
<tr><td>Present</td><td>7068</td></tr>
<tr><td>Future</td><td>1345</td></tr>
<tr><td>Unspecified</td><td>1608</td></tr>
<tr><td rowspan="2"># Event Triggers (by Polarity)</td><td>Positive</td><td>26276</td></tr>
<tr><td>Negative</td><td>2106</td></tr>
<tr><td rowspan="2"># Event Triggers (by Modality)</td><td>Asserted</td><td>25671</td></tr>
<tr><td>Other</td><td>2717</td></tr>
<tr><td colspan="2"># Arguments</td><td>49404</td></tr>
</table>

In Figure 7, you can see a graph of event types and subtypes, in which the size of subtypes correspond to their frequency in the gold-standard dataset. According to this figure, the most frequent event types are “*meeting*”, “*death*” and “*homicide*”.

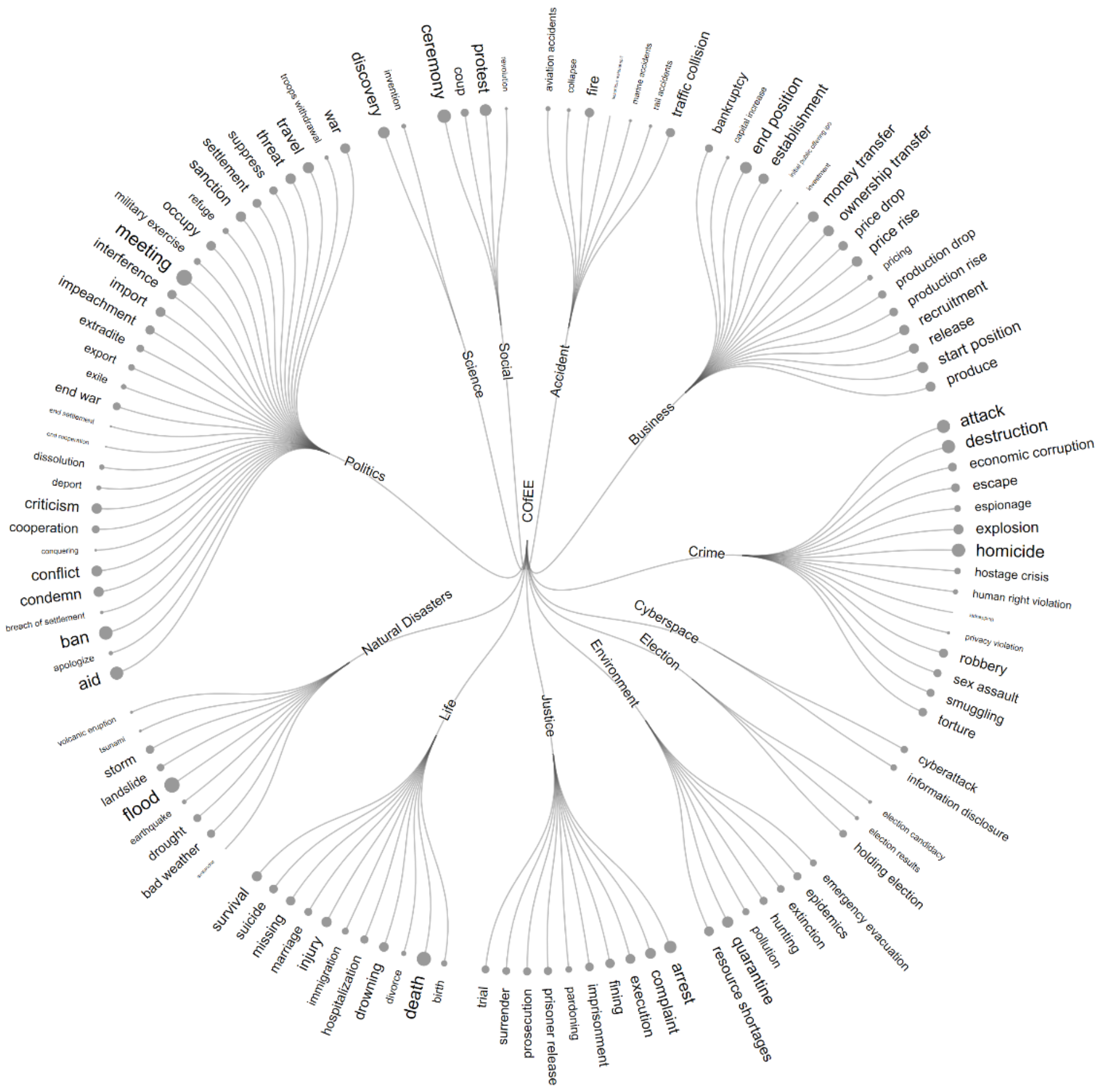

*Figure 7. The graph of the gold-standard dataset’s event types and subtypes. The size of each event subtype is an indication of the frequency of its occurrence in the dataset.*

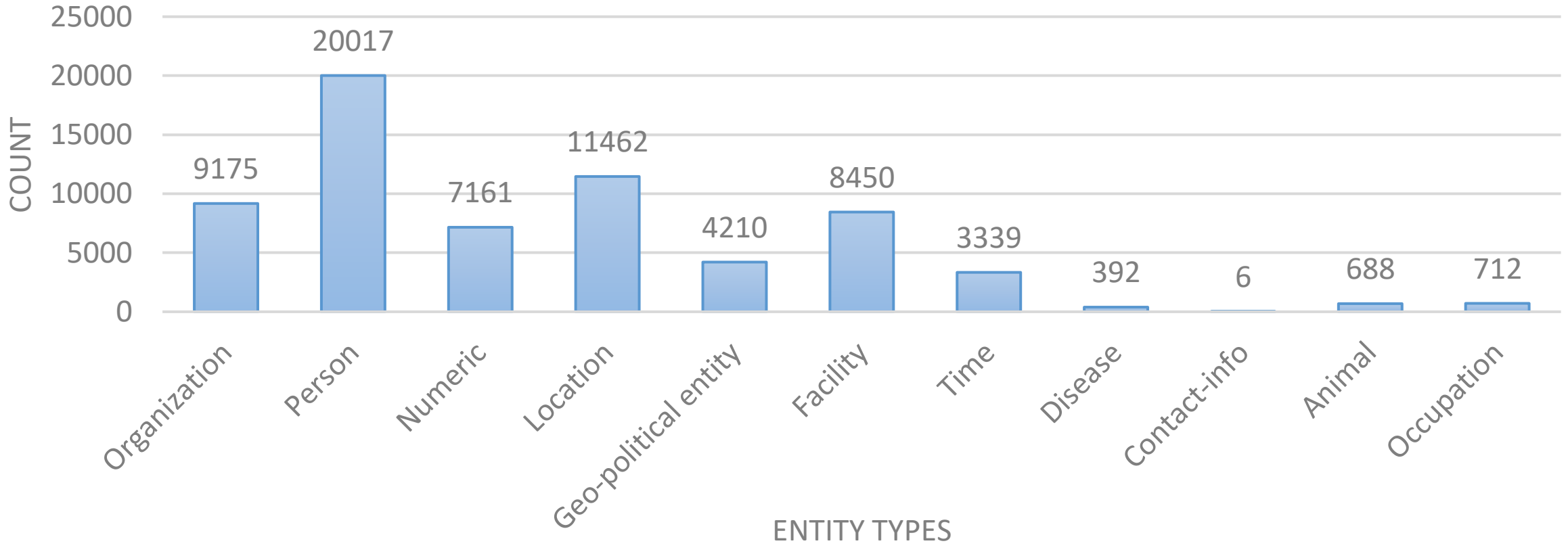

*Figure 8. Distribution of entity types in the gold-standard data*

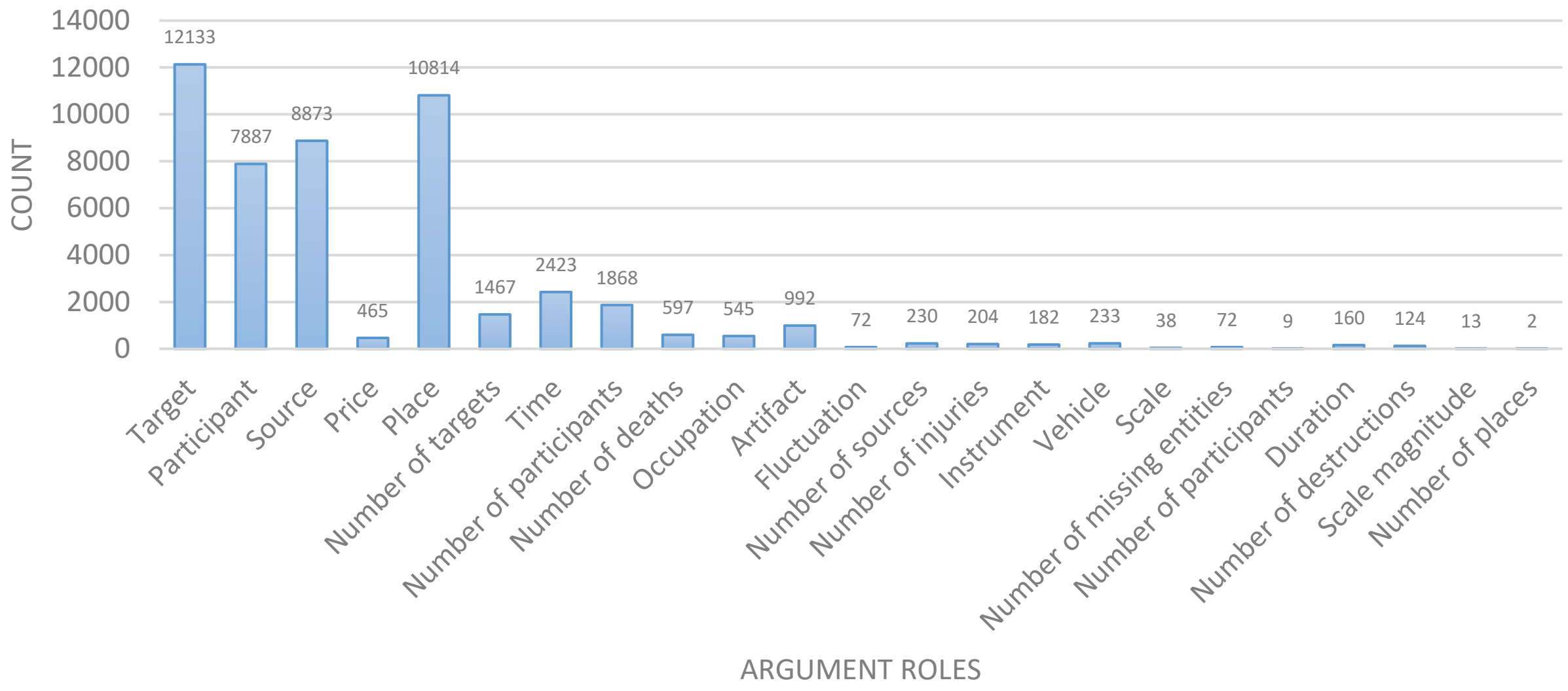

*Figure 9. Distribution of argument roles in the gold-standard data*

Figure 8 and 9 show respectively, the distribution of entity types and argument roles in the gold-standard data. The most frequent entity types in this dataset are "*person*", "*location*" and "*organization*". Also, the most frequent argument roles are "*target*", "*source*" and "*place*".

## 6: SMEE Method

In this section, we propose a Supervised Method for Event Extraction, named *SMEE*, based on deep learning that can be considered as a baseline. This method predicts both event triggers and their arguments for a given input sentence. In Figure 10, the architecture of the proposed method for event extraction is depicted. As the argument identification and the argument role classification tasks have the same architecture in our method (except in the last layer, i.e., the output of the method), we present them together for the sake of simplicity.

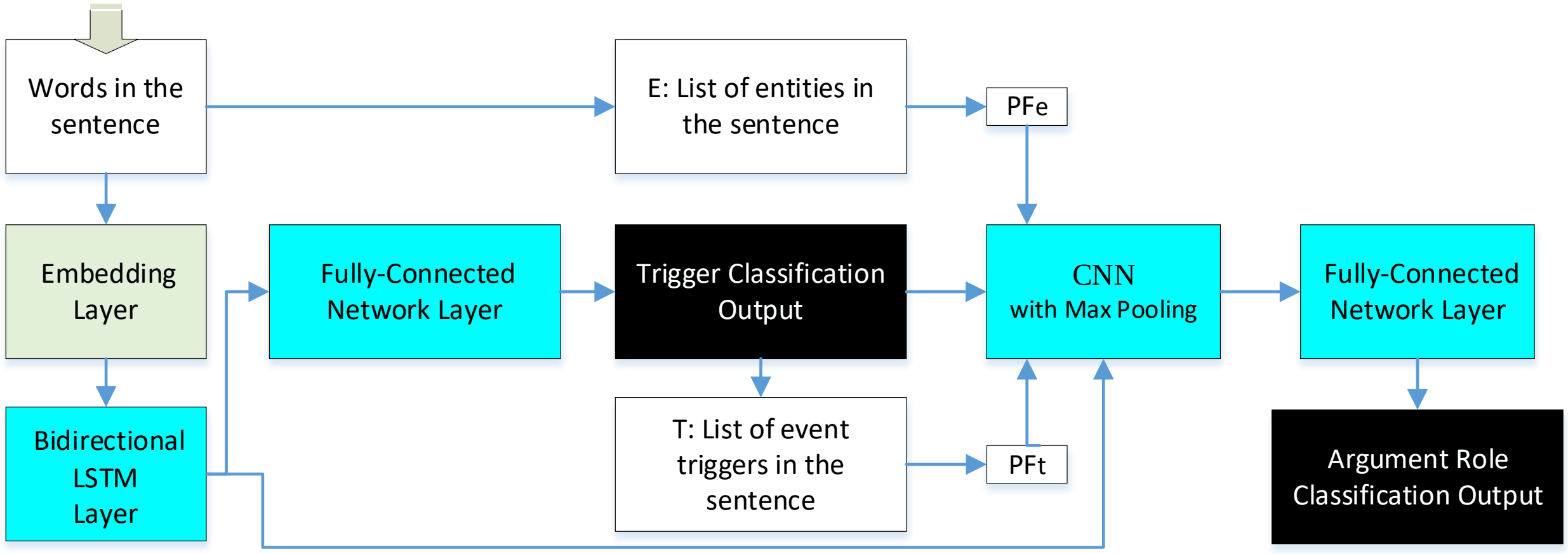

*Figure 10. Architecture of the SMEE method for event extraction*

Given a sentence S = [$w_1$, $w_2$, …, $w_n$] with length *n*, each word w is represented by a real-valued vector that is the result of the embedding layer. We use GloVe word embedding [60] to create a vector representation of the words. GloVe is trained on more than one million news in Persian language. Similar to Liu et al. [45, 46], by padding shorter sentences and cutting off longer ones, the maximum length L=50 is considered in the experiments. To determine the word context, the corresponding vectors are passed through a bidirectional LSTM layer [61]. Finally, the output of the bidirectional layer of the LSTM is fed into a fully-connected network to identify event triggers and subtypes associated with each word.

To detect the argument roles according to each event trigger *t* (detected by the trigger classification task) and entity *e* (extracted from the gold-standard dataset), the vector generated by the bidirectional LSTM, the output vector of the event subtype classification and the position feature (PF) are concatenated. $PF_{tk}$ and $PF_{ek}$ specify the relative distance between $k^{th}$ word of a given sentence, with the trigger candidates *t* and the argument candidate *e* respectively. *PF* is necessary to specify which words are the trigger and argument candidate [55, 62, 63]. To encode the position feature (PF), each distance value is represented by an embedding vector which is generated randomly. Then, these concatenated vectors are passed into the Convolutional Neural Network (CNN) with max pooling to automatically extract the most important features of the different parts of the sentence. Finally, the output vectors are passed into a fully-connected network layer where the output layer predicts the role of each argument candidate for each event trigger.

There are 12 entity types, 119 event subtypes and 21 argument roles in the COfEE event ontology. The purpose of the event trigger classification task is to classify each word of a given sentence into 239 categories in IOB annotation schema according to COfEE (along with the default "NoEvent" category). Also, the goal of the argument role classification task is to classify the role of each argument for each trigger into 22 categories (along with the default "NoRole" category).

## 7: Experimental Results

In the section, the generalization of the COfEE ontology is examined, and then the SMEE method is evaluated.

## 7-1: Evaluation of COfEE

The Wikipedia event portal[22] lists the most important news stories (containing events) published in more than two thousand online news agencies over time. In order to determine the coverage and generality of our EO we have performed an experiment on the corpus of these Wikipedia events over the past 10 years (around 55k news records). Because Wikipedia is a crowd-sourcing platform, these events are listed by different people with different points of views; this makes it less susceptible to personal bias. The corpus includes events from more than 10 general categories[23], which is, in our opinion, diverse enough. For every event subtype of COfEE, we have considered several representing keywords, listed on Table 9. These keywords are chosen following

[22] https://en.wikipedia.org/wiki/Portal:Current_events

[23] Arts and culture, armed conflicts and attacks, business and economy, disasters and accidents, health and environment, international relations, law and crime, politics and elections, science and technology, sports

four steps: (1) we prepared an initial seed by human experts representing the predefined subtypes. (2) We then fed the initial seed to Google News 300M Word2Vec [64] model to find 10 most similar words for each seed word. (3) Then, we combined the result of the last two steps and found the lexeme of the resulting words. (4) Finally human experts revised the list of words to correct the errors occurred during the previous steps.

The keywords that represent the events are carefully chosen to be both **collector** and **blocker**. Hence, the chosen keywords express the exact events they are meant to represent. This is a crucial step. One ambiguous keyword may include events that do not belong to the one it should, whereas a very specific keyword may not cover all the forms in which an event could be expressed.

In choosing the representing keywords, their most common meanings in the news have been taken into account e.g., the word "fine" is interpreted as "money extracted as a penalty" instead of "being satisfactory". Moreover, a single keyword may be associated with more than one event subtype, for instance the term "capture" may appear in both "Conquering" and "Arrest" event subtypes.

In Figure 11, a graph of these event types and subtypes is shown, in which the subtypes' size is proportional to their term frequency in Wikipedia corpus. The exact numbers are reported in Table 9. "*Death*", "*Attack*" and "*Protest*" are the most frequently seen event subtypes. "*Election*" is another quite frequent and controversial event subtype that is commonly covered in the media. "*Epidemics*", on the other hand, have recently had a high impact on our society due to COVID-19, justifying why they are a prominent subtype of events.

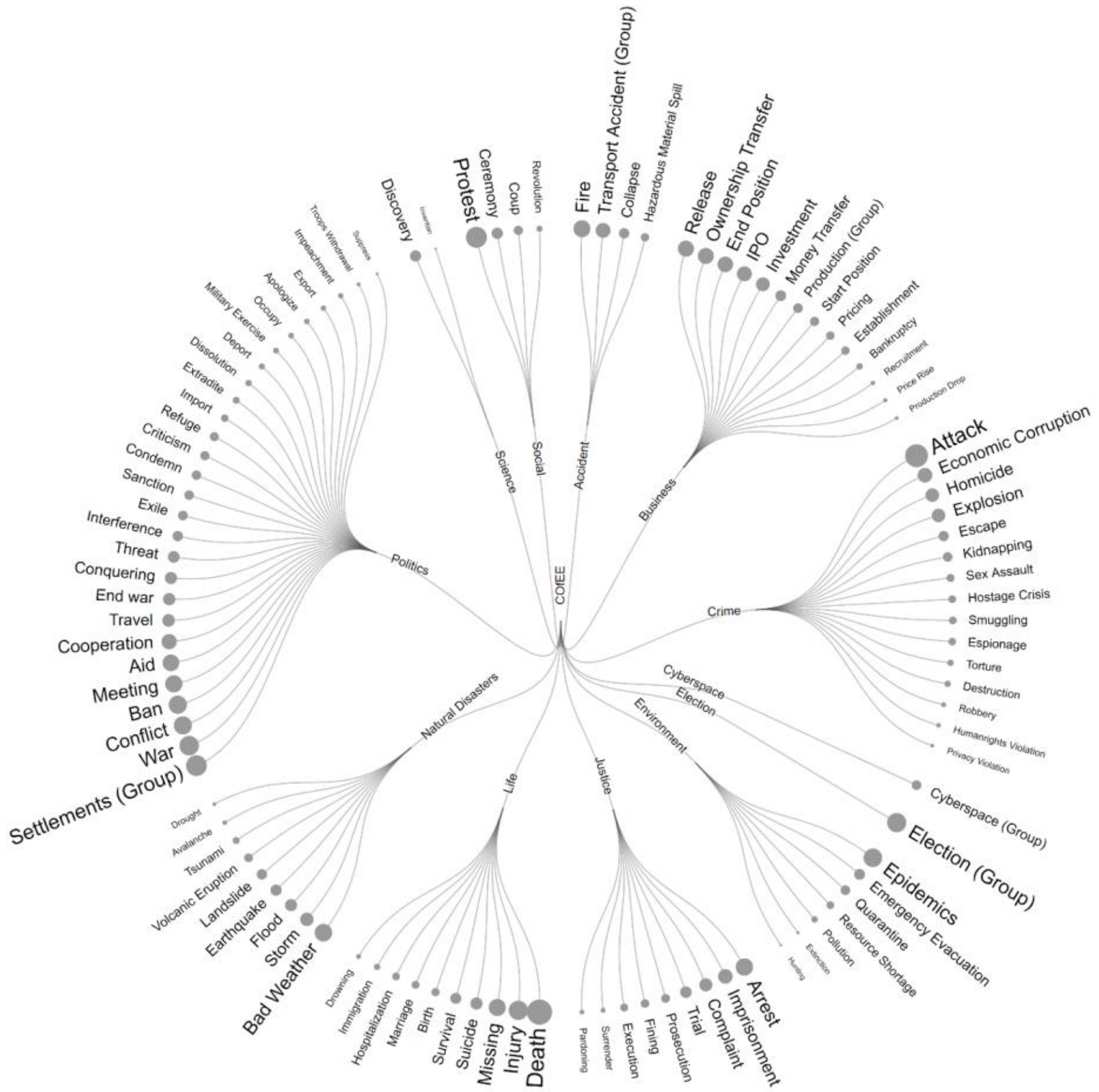

*Figure 11. Graph of COfEE's events. The size of each event subtype shows the frequency of its occurrence in the Wikipedia corpus. Subtype names that are followed by (Group) denote aggregated results for the corresponding event type.*

Several common event types are discussed in the following lines. Life's most common subtypes are "*Death*", "*Injury*" and "*Missing*". "*Crime*" is mainly composed of "*Attack*", "*Economic Corruption*" and "*Homicide*". "*Settlements*", "*Wars*", "*Conflicts*", "*Ban*", and "*Meetings*" are the most seen subtypes in "*Politics*" events. In "*Natural Disasters*", the most common subtype is "*Bad Weather*". It is "*Arrest*" that takes the most share of "*Justice*". "*Fire*" and "*Transport Accident*" are the most appearing "*Accident*" subtypes. "*Release*", "*Ownership Transfer*" and "*End Position*" mostly make up "*Business*" events. "*Social*" events are mostly related to "*Protest*". "*Environment*" consists primarily of "*Epidemics*".

Interestingly, COfEE has been found to cover more than 88% of the Wikipedia corpus, denoting a high generalization potential of the ontology. The coverage has been calculated as the proportion of number of Wikipedia records that contains at least one of the events in COfEE to the number of all records in the corpus. It should be noted that the total number of events in the corpus is expected to be higher than those detected by COfEE, due to the fact that the focus of the proposed framework is identifying events with more frequency and those requiring instant monitoring.

## 7-2: SMEE Experimental Results

Same as in the previous works [45, 46, 53-56], we judge the correctness of event extraction based on the following criteria: A trigger is correctly classified if its event subtype and offsets match those of a reference trigger. An argument is correctly identified if its event subtype and offsets match those of any of the reference argument mentions. An argument is correctly classified if its event sub-type, offsets and argument role match those of any of the reference argument mentions. To this regard we use Precision (*P*), Recall (*R*) and F-measure (*F1*) as the evaluation metrics.

For the test set, we randomly select 10-20% of the gold-standard data for each event subtype, so that low-occurring events are not dominated by more frequent ones. Finally, there are 20,812 sentences in the training set and 3307 sentences in the test set.

The method is implemented using Keras Deep Learning library and is trained on Google Colab using a Tesla K80 GPU. Table 6, shows the hyperparameters of the proposed method tuned on the validation set through grid search. For training the parameters, cross entropy is selected as the loss function with equal weights for the two output models. We use Stochastic Gradient Descent with shuffled mini-batches and Adam update rule [65] to minimize the loss.

*Table 6. Hyperparameters in SMEE*

| Hyperparameters | |
|---|---|
| GloVe embedding size | 200 |
| Positional embedding size (PF) | 10 |
| LSTM hidden units | 64 |
| Dropout fraction | 0.5 |
| Batch size | 32 |
| The number of filters in CNN | 64 |
| The convolution window size in CNN | 3 |
| Number of hidden units (in the event subtype classification output) | 384 |
| Number of hidden units (in the argument identification output) | 128 |
| Number of hidden units (in the argument role classification output) | 128 |

Table 7 shows the overall effectiveness of the proposed methods on the test set. In this table, three versions of *SMEE* are proposed, *SMEE* with LSTM and CNN layers ("SMEE +LSTM +CNN"), *SMEE* with LSTM layer and without CNN layer ("SMEE +LSTM -CNN") and finally, *SMEE* without LSTM layer and with CNN layer ("SMEE -LSTM +CNN"). From the results, we can see that the "SMEE +LSTM +CNN" model achieves the best effectiveness in argument identification and argument role classification tasks among the methods according to F1-score. Also, the best result is achieved by "SMEE +LSTM -CNN" in event trigger classification.

LSTM layer improves F1-score by 6.3% and 6.2% for the trigger classification task and argument role classification task, respectively. Additionally, CNN layer improves F1-score by 13% for the argument role classification task.

Table 7: Overall effectiveness of the proposed method.

| Method | Event trigger Classification (%) | | | Argument Identification (%) | | | Argument Role Classification (%) | | |
|---|---|---|---|---|---|---|---|---|---|
| | P | R | F1 | P | R | F1 | P | R | F1 |
| *SMEE +LSTM +CNN* | 80.1 | 73.4 | 76.6 | **63.3** | **53.2** | **57.8** | **57.9** | **48.6** | **52.8** |
| *SMEE -LSTM +CNN* | 73.5 | 67.4 | 70.3 | 55.1 | 51 | 53 | 48.5 | 44.9 | 46.6 |
| *SMEE +LSTM -CNN* | **84.4** | **78.5** | **81.4** | 62.5 | 42.5 | 50.6 | 49.2 | 33.5 | 39.8 |

## 8: Conclusion

In this paper, we introduced an event ontology, "COfEE", for extracting events from unstructured text which cover many contemporary issues. This ontology represents events in a structure which includes event type/subtype and argument roles. COfEE includes 12 event types, 119 event subtypes and 21 argument roles that are carefully chosen based on the domain knowledge of human experts and data-driven observations. To study the coverage and generality of the proposed ontology, an experiment on Wikipedia event portal is devised. It is shown that the ontology covers a wide range of various events. To facilitate the process of preparing gold-standard data according to COfEE for the training of machine learning methods, we propose a free, language-independent and multi-user tool to annotate texts which can boost the speed of user annotation process.

We also introduced the first gold-standard data for event extraction in Persian language based on the COfEE event ontology, annotated by 10 human experts. The prepared gold-standard data includes 24k sentences include 65K entities, 28K event triggers and 49K arguments. Moreover, we proposed a supervised-learning method to automatically extract events from the aforementioned dataset. It was shown that the proposed method achieved 84.4%, 78.5% and 81.4% on Precision, Recall and F1-score respectively, in trigger classification task and 57.9%, 48.6% and 52.8% in argument role classification task. In future, we plan to further develop the proposed machine learning method to improve the effectiveness of the event extraction task in Persian. Also, we intend to generate gold-standard data based on COfEE for some other languages, including English and Arabic.

*Table 8. Guidelines for introducing argument roles in different event subtypes in COfEE (bold-words=event triggers, underlined-words= arguments, superscript-numbers=the role of the argument in a specific event trigger). In front of the argument roles, the entity classes are listed in parenthesis which can be used for the corresponding roles). For simplicity, the event subtypes with the same roles are grouped into one row. The news texts are gathered from different online news agencies.*

## 1) Event type: Life

**Event subtypes (E):** Death[1-1], Injury[1-2], Birth[1-3], Drowning[1-4], Survival[1-5], Marriage[1-6], Divorce[1-7], Hospitalization[1-8], Missing[1-9]

**Argument roles (R):** Participant[1] (Person, Animal), Time[4] (Time), Place[5] (GP, Location), Number of Participants[13] (Numeric)

- Around 40[E1-1(R13)] fishermen[E1-1(R1)] in Zambia[E1-1(R5)] are ***dead***[E1-1].
- Cristiano Ronaldo [E1-2(R1)] appeared to suffer a muscle ***injury*** [E1-2] on Monday [E1-2(R4)] in Portugal [E1-2(R5)].
- Jackson[E1-3(R1)] was ***born***[E1-3] on August 29, 1958[E1-3(R4)], in Gary, Indiana[E1-3(R5)].
- A 12-year-old Somali girl[E1-4(R1)] ***drown***[E1-4] whilst playing in a river with her friends in Manchester, England[E1-4(R5)].
- Twelve[E1-5(R13)] people[E1-5(R1)] were ***rescued*** [E1-5] from the sea off Oinousses[E1-5(R5)].
- Putin[E1-6(R1), E1-7(R1)] was ***married*** [E1-6] to Lyudmila Putina [E1-6(R1), E1-7(R1)] from 1983[E1-6(R4)] until their ***divorce*** [E1-7], announced in 2013[E1-7(R4)].
- Palestinian President Mahmoud Abbas [E1-8(R1)] was ***hospitalized*** [E1-8] in the West Bank [E1-8(R5)] on Sunday [E1-8(R4)].
- A four-year-old boy [E1-9(R1)] who had gone ***missing*** [E1-9] Saturday [E1-9(R4)] in Mackenzie, B.C [E1-9(R5)] when he became separated from his mother.

**Event subtypes (E):** Immigration[1-10]

**Argument roles (R):** Source[2] (Person, Animal), Target[3] (GP, Location), Time[4] (Time), Place[5] (GP, Location), Number of Sources[18] (Numeric)

- More than 250,000 [E1-10(R18)] Germans [E1-10(R2)] ***emigrate*** [E1-10] from Germany [E1-10(R5)] each year [E1-10(R4)], according to German government statistics.

**Event subtypes (E):** Suicide[1-11]

**Argument roles (R):** Source[2] (Person, Animal), Time[4] (Time), Place[5] (GP, Location), Instrument[6] (Facility), Number of Sources[18] (Numeric)

- In 2005 [E1-11(R4)], South Korean actress Lee Eun Joo [E1-11(R2)] committed ***suicide*** [E1-11] at the age of 24.

## 2) Event type: Natural disaster

**Event subtype (E):** Volcanic eruption[2-1], Tsunami[2-2], Earthquake[2-3], Landslide[2-4], Avalanche[2-5], Storm[2-6], Flood[2-7], Drought[2-8], Bad weather[2-9]

**Argument roles (R):** Participant[1] (Person, Animal, Facility), Time[4] (Time), Place[5] (GP, Location), Scale (Magnitude)[20] (Numeric), Number of Injuries[14] (Numeric), Number of Deaths[15] (Numeric), Number of Missing Entities[16] (Numeric), Number of Destructions[17] (Numeric), Number of Participants[13] (Numeric)

- Around 110 [E2-6(R15), E1-1(R13)] people [E2-6(R1), E1-1(R1)] are ***killed*** [E1-1] in a ***dust storm*** [E2-6] in northwestern India [E2-6(R5), E1-1(R5)].
- ***Heavy snow***[E2-9] in northern Iran [E2-9(R5), E3-6(R5)] has left around 480,000 [E2-9(R17), E3-6(R13)] homes [E2-9(R1), E3-6(R1)] ***without power*** [E3-6].
- Magnitude 5 [E2-3(R20)] **quake** [E2-3] in southwestern China [E2-3(R5), E1-1(R5), E1-2(R5)] ***kills*** [E1-1] 4 [E2-3(R15), E1-1(R13)], ***injures*** [E1-2] 23 [E2-3(R14), E1-2(R13)].

## 3) Event type: Environment

**Event subtype (E):** Pollution[3-1], Epidemics[3-2]

**Argument roles (R):** Source[2] (GP, Person, Animal, Org, Facility, Disease), Target[3] (Person, Animal, Org), Time[4] (Time), Place[5] (GP, Location), Number of Targets[19] (Numeric), Number of Injuries[14] (Numeric), Number of Deaths[15] (Numeric),

- In 2015 [E1-1(R4), E3-1(R4)], more than 1.1 million [E1-1(R13), E3-1(R19)] people [E1-1(R1), E3-1(R3)] in China [E1-1(R5), E3-1(R5)] were estimated to have ***died***[E1-1] from ***air pollution*** [E3-1].
- According to the International Committee of the Red Cross, a cholera [E3-2(R2)] ***outbreak*** [E3-2] has since April 27 [E3-2(R4),E1-1(R4),E1-2(R4)] ***killed*** [E1-1] at least 115 [E3-2(R15), E1-1(R13)] people [E3-2(R3), E1-1(R1)] and left another, [E3-2(R3), E1-2(R1)] 8,500 [E3-2(R14), E1-2(R13)] ***ill*** [E1-2] across Yemen [E3-2(R5), E1-1(R5), E1-2(R5)].

- About 8,600 [E3-2(R19)] people [E3-2(R3)] have been diagnosed with coronavirus [E3-2(R2)] ***infections*** [E3-2] in California [E3-2(R5)].

**Event subtype (E):** Hunting[3-3],

**Argument roles (R):** Source[2] (Person, Animal, Org), Target[3] (Animal), Time[4] (Time), Place[5] (GP, Location), Number of Sources[18] (Numeric), Number of Targets[19] (Numeric)

- Botswana auctions licenses to ***hunt*** [E3-3] 70 [E3-3(R19)] elephants [E3-3(R3)] in effort to reduce conflict with Persons.

**Event subtype (E):** Extinction[3-4], Emergency Evacuation[3-5], Resource shortages [3-6]

**Argument roles (R):** Participant[1] (Person, Animal), Time[4] (Time), Place[5] (GP, Location), Number of Participants[13] (Numeric)

- The South China Tiger[E3-4(R1)] is thought to *be* ***extinct***[E3-4] in the wild [E3-4(R5)] as it hasn't been spotted since the 1970s [E3-4(R4)].
- The Javan Rhino [E3-4(R1)] is the most threatened with ***extinction*** [E3-4] with the total population of only 60 [E3-4(R13)] surviving.
- Over one million[E3-5(R13), E2-9(R13)] people[E3-5(R1), E2-9(R1)] in the Japanese prefectures of Miyazaki and Kagoshima [E3-5(R5), E2-9(R5)] are ordered to ***evacuate*** [E3-5] their homes due to ***torrential rainfall*** [E2-9].
- The death toll in the U.S. [E2-6(R5)] attributed to ***Hurricane Sandy*** [E2-6] rises to at least 90 [E2-6(R15)], as millions [E3-6(R13), E2-6(R13)] of people[E3-6(R1), E2-6(R1)] in the Northeastern United States[E3-6(R5), E2-6(R5)] continue to deal with ***power outages*** [E3-6], ***gasoline shortages*** [E3-6] and ***sparse public transportation***[E3-6].
- Crimea[E3-6(R5)] is ***without power*** [E3-6] after transmission towers in Ukraine's Kherson Oblast[E4-2(R5)] were ***blown up*** [E4-2] by unknown people [E4-2(R2)].
- ***The storm*** [E2-6] causes ***power outages*** [E3-6] to 550,000 [E3-6(R13), E2-6(R13)] customers [E3-6(R1), E2-6(R1)] in Florida [E2-6(R5), E3-6(R5), E3-5 (R5)], Georgia [E2-6(R5), E3-6(R5), E3-5(R5)] , and Alabama [E2-6(R5), E3-6(R5), E3-5 (R5)] , and 375,000 [E3-5(R13), E2-6(R13)] people [E3-5(R1), E2-6(R1)] have been ordered to ***evacuate*** [E3-5].
- An estimated 1.25 million [E3-6(R13)] people [E3-6(R1)] in South Sudan [E3-6(R5)] are on **the brink of *starvation*** [E3-6], according to the latest food and security analysis update.

**Event subtype (E):** Quarantine[3-7]

**Argument roles (R):** Participant[1] (Person, Org, Animal), Time[4] (Time), Place[5] (GP, Location), Duration[12] (Time), Number of Participants[13] (Numeric)

- King Felipe VI [E3-7(R1)] voluntarily ***quarantines*** [E3-7] for 10 days [E3-7(R12)] after being in close contact with someone who later tested positive for COVID-19.
- Mateusz Morawiecki [E3-7(R1), E7-10(R1)] goes into ***quarantine*** [E3-7] after ***making contact*** [E7-10] with a person [E7-10(R1)] who has tested positive for COVID-19.

## 4) Event type: Crime

**Event subtype (E):** Attack[4-1], Explosion[4-2], Hostage Crisis[4-3]

**Argument roles (R):** Source[2] (GP, Person, Org, Animal), Target[3] (Person, Animal, Org, Facility), Instrument[6] (Facility), Time[4] (Time), Place[5] (GP, Location), Number of Injuries[14] (Numeric), Number of Deaths[15] (Numeric), Number of Destructions[17] (Numeric), Number of Sources[18] (Numeric), Number of Targets[19] (Numeric)

- A man [E4-1(R2)] with a knife[E4-1(R6)] ***attack***[E4-1] leaves one [E4-1(R15), E1-1(R13)] ***dead*** [E1-1] and three [E4-1(R14), E1-2(R13)] ***injured*** [E1-2] at the University of Texas at Austin[E4-1(R5), E1-1(R5), E1-2(R5)].
- ***The explosion*** [E4-2] of a roadside bomb[E4-2(R6)] ***kills*** [E1-1] at least 11 [E4-2(R15), E1-1(R13)] people [E4-2(R3), E1-1(R1)], all from the same family, in Logar Province [E4-2(R5), E1-1(R5)].
- Gunman[E1-1(R1), E4-3(R2)] ***killed***[E1-1] after taking two [E4-3(R19)] women[E4-3(R3)] ***hostage*** [E4-3] at UPS facility in New Jersey [E1-1(R5), E4-3(R5)].

**Event subtype (E):** Escape[4-4]

**Argument roles (R):** Source[2] (Person, Animal), Time[4] (Time), Place[5] (GP, Location), Number of Sources[18] (Numeric), Vehicle[7] (Facility)

- Two [E4-4(R18)] youth [E4-4(R2)] ***escaped*** [E4-4] from the Woodside Juvenile Rehabilitation Center's temporary location in St. Albans [E4-4(R5)], according to state officials.

**Event subtype (E):** Smuggling[4-5]

**Argument roles (R):** Source[2] (GP, Person, Org), Target[3] (GP, Location), Time[4] (Time), Place[5] (GP, Location), Artifact[8] (Person, Animal, Facility), Vehicle[7] (Facility)

Iran's Guards [E5-2(R2)] have ***seized*** [E5-2] a vessel [E5-2(R3), E4-5(R7)] in the Gulf [E5-2(R5)] for allegedly ***smuggling***[E4-5] 250,000 litres of diesel fuel[E4-5(R8)] to the United Arab Emirates [E4-5(R3)].

**Event subtype (E):** Destruction[4-6], Sex assault[4-7], Kidnapping[4-8], Homicide[4-9], Torture[4-10]

**Argument roles (R):** Source[2] (Person, Org, Facility), Target[3] (Person, Facility), Instrument[6] (Facility), Time[4] (Time), Place[5] (GP, Location), Number of Sources[18] (Numeric), Number of Targets[19] (Numeric)

- ISIS [E4-6(R2)] posts a video showing ***the destruction*** [E4-6] of Mosul Museum [E4-6(R3)], the second largest in Iraq [E4-6(R5)] and rich in artifacts from thousands of years of Iraqi history.
- Former U.S. Congressman Anthony Weiner[E4-7(R2)] pleads guilty to ***sexting***[E4-7] with a 15 year old girl[E4-7(R3)].
- Daisy McCrackin[E4-8(R3)] and Joseph Capone [E4-8(R3)] were allegedly ***kidnapped***[E4-8] and held for ransom in Compton [E4-8(R5)] last year [E4-8(R4)].
- A 22-year-old man [E4-8(R2)] and a 23-year-old woman [E4-8(R2)] reportedly ***kidnapped*** [E4-8] a 2-year-old boy [E4-8(R3)] from his adoptive mother.
- Sacramento Mother [E4-9(R2)] Suspected Of ***Homicide*** [E4-9] after ***Death*** [E1-1] of Her Young Son [E4-9(R3), E1-1(R1)].
- Taliban insurgency Gunmen [E4-9(R2)] ***assassinate*** [E4-9] Atiqullah Rawoofi [E4-9(R3)], the head of the Secretariat of the Supreme Court of Afghanistan, in the outskirts of Kabul [E4-9(R5)].
- The Abu Dhabi Police[E5-2(R2)] ***arrested***[5-2] a mother [E5-2(R3), E4-10(R2)] in the emirate of Abu Dhabi [E5-2(R5), E4-10(R5)] for ***torturing***[E4-10] her daughter [E4-10(R3)].

**Event subtype (E):** Robbery[4-11], Economic corruption[4-12]

**Argument roles (R):** Source[2] (Person, Org, Animal), Target[3] (Person, Animal, Org), Time[4] (Time), Place[5] (GP, Location), Artifact[8] (Animal, Facility), Price[21] (Numeric), Number of Sources[18] (Numeric), Number of Targets[19] (Numeric)

- Three[E4-11(R18)] men[E4-11(R2)] ***stealing***[E4-11] a safe full of jewellery [E4-11(R8)] from a store [E4-11(R3)] in Sydney's inner city [E4-11(R5)].
- A 54 year old Montana woman [E4-12(R2)] has pleaded guilty to ***embezzling*** [E4-12] more than $ 630,000 [E4-12(R21)] from customers of her employer [E4-12(R3)] in 2016 [E4-12(R4)].
- Najva Lasheidaei [E4-12(R2)] and her husband Vahid Behzadi [E4-12(R2)] are said to have ***smuggled*** [E4-12] hard currency [E4-12(R8)] and ***laundered*** [E4-12] $200 million [E4-12(R21)].
- A federal jury [E5-10(R2)] in Little Rock [E5-10(R5)] ruled that Walmart [E5-10(R3), E4-11(R2)] ***stole*** [E4-11] trade secrets [E4-11(R8)] from Agtech company Zest Labs [E4-11(R3)], a breach that came with a $115 million [E5-10(R21)] ***fine*** [E5-10].

**Event subtype (E):** Espionage[4-13], Copyright violation[4-14], Human right violation[4-15], Privacy violation[4-16]

**Argument roles (R):** Source[2] (GP, Person, Org, Facility), Target[3] (GP, Person, Org, Facility), Time[4] (Time), Place[5] (GP, Location)

- The Delhi police [E5-2(R2)] said that journalist Rajeev Sharma [E5-2(R3), E4-13(R2)], who was ***arrested***[E5-2] on charges of ***spying*** [E4-13] two days ago [E5-2(R4)], was allegedly passing on sensitive information to Chinese intelligence agencies.
- The Mumbai Police[E5-2(R2)] on Tuesday [E5-2(R4)] ***arrested*** [E5-2] Maha Movie television channel's CEO Sanjay Verma [E5-2(R3), E4-14(R2)] in an alleged ***copyright violation*** [E4-14] case, a police official said.
- The Trump administration [E7-12(R2)] imposes additional ***sanctions***[E7-12] against Belarusian officials [E7-12(R3), E4-15(R2)] and four offices [E7-12(R3), E4-15(R2)] for ***human rights abuses*** [E4-15] that occurred during ***protests***[E8-1] against President Alexander Lukashenko [E8-1(R3)].
- Facebook's Onavo security app[E4-16(R2)] taken down from Apple store for ***violating user privacy*** [E4-16].

## 5) Event type: Justice

**Event Subtype (E):** Complaint[5-1], Arrest[5-2], Pardoning[5-3], Prosecution[5-4], Execution[5-5], Trial[5-6], Surrender[5-7], Prisoner Release[5-8]

**Argument roles (R):** Source[2] (GP, Org, Person), Target[3] (GP, Person, Org, Facility), Time[4] (Time), Place[5] (GP, Location), Number of Targets[19] (Numeric)

1. People [E5-1(R2)] are ***complaining*** [E5-1] to apple [E5-1(R3)] for update of new iPhone is filled with bugs.
2. West Midlands Police [E5-2(R2)] ***arrest*** [E5-2] five [E5-2(R19)] men [E5-2(R3)] suspected of membership of National Action.
3. President Donald Trump [E5-3(R2)] announced that he [E5-3(R2)] ***pardoned*** [E5-3] Eddie DeBartolo Jr. [E5-3(R3)] and Bernie Kerik [E5-3(R3)].
4. The American woman [E5-4(R3), E4-9(R2)] accused of ***killing*** [E4-9] teenager Harry Dunn [E4-9(R3)] is now ***wanted*** [E5-4] by Interpol [E5-4(R2)].
5. Saudi Arabia [E5-5(R2)] ***executes*** [E5-5] Yemeni man [E5-5(R3), E4-1(R2)] who ***attacked*** [E4-1] Spanish performers [E4-1(R3)].
6. A judge [E5-6(R2)] rules that former President of Guatemala Alfonso Portillo [E5-6(R3)] must stand ***trial*** [E5-6].
7. Canadian singer Justin Bieber [E5-7(R2)] ***surrenders*** [E5-7] to Toronto Police [E5-7(R3)] to face assault charges.

8. More than 1,000 [E5-8(R19)] Illinois prisoners [E5-8(R3)] to be ***released*** [E5-8] under COVID-19 lawsuit settlement.
9. Alex Saab, the Venezuelan diplomat [E5-2(R3), E5-8(R3), E4-12(R2)] ***arrested*** [E5-2] in Cape Verde [E5-2(R5)] in June 2020 [E5-2(R4)] on the orders of the US [E5-2(R2)] over allegations of ***corruption*** [E4-12] has been ***released from prison*** **custody** [E5-8] on the orders of the ECOWAS court [E5-8(R2)].

**Event subtype (E):** Imprisonment[5-9]

**Argument roles (R):** Source[2] (GP, Org, Person), Target[3] (Person), Time[4] (Time), Place[5] (GP, Location), Duration[12] (Time), Number of Targets[19] (Numeric)

- Man [E5-9(R3), E4-7(R2)] ***gets prison*** [E5-9] in ***sex assault*** [E4-7] of 8-year-old [E4-7(R3)].
- A man [E5-9(R3), E4-8(R2), E4-9(R2)] is sentenced to 25 years [E5-9(R12)] ***imprisonment***[E5-9] for the ***kidnapping*** [E4-8] and ***Murder*** [E4-9] of Anni Dewani [E4-8(R3), E4-9(R3)] in South Africa [E4-8(R5), E4-9(R5), E5-9(R5)].

**Event subtype (E):** Fining [5-10]

**Argument roles (R):** Source[2] (GP, Org, Person), Target[3] (GP, Org, Person), Time[4] (Time), Place[5] (GP, Location), Price[21] (Numeric)

- U.S. car maker General Motors [E5-10(R3), E6-4(R2)] ***pays*** [E6-4] $900 million [E5-10(R21), E6-4(R21)] ***fine***[E5-10] to settle a criminal lawsuit over problems with the ignition system in its small cars.

## 6) Event type: Business

**Event subtype (E):** Start-Position [6-1], Recruitment[6-2], End-Position[6-3]

**Argument roles (R):** Source[2] (GP, Org, Person), Target[3] (Person), Time[4] (Time), Place[5] (GP, Location), Occupation[9] (Occupation), Contact-info[11] (Contact-info), Number of Targets[19] (Numeric)

- On July 16, 2012 [E6-1(R4)], Mayer[E6-1(R3)] was ***appointed*** [E6-1] president [E6-1(R9)] and CEO [E6-1(R9)] of Yahoo! [E6-1(R2)].
- SBPC [E6-2(R2)] is in ***need*** [E6-2] of two [E6-2(R19)] special person [E6-2(R3)] to serve as our Transportation Specialist [E6-2(R9)], For more info, please contact Cindy at 858-509-2587 [E6-2(R11)] or cindy@solanapres.org [E6-2(R11)].
- U.S. President Donald Trump[E6-3(R2)] ***removes***[E6-3] James Comey [E6-3(R3)] as Director of the Federal Bureau of Investigation [E6-3(R9)], with Andrew G. McCabestepping [E6-3(R3)] in as acting director [E6-3(R9)].

**Event subtype (E):** Money transfer[6-4]

**Argument roles (R):** Source[2] (GP, Person, Org), Target[3] (GP, Person, Org), Time[4] (Time), Place[5] (GP, Location), Price[21] (Numeric)

- Mexico [E6-4(R2)] will ***pay***[E6-4] $ 159.88 million [E6-4(R21)] to the World Health Organization [E6-4(R3)] to secure vaccine access through the agency's COVAX plan.

**Event subtype (E):** Ownership transfer[6-5]

**Argument roles (R):** Source[2] (GP, Person, Org), Target[3] (GP, Person, Org), Time[4] (Time), Place[5] (GP, Location), Artifact[8] (Animal, Facility), Price[21] (Numeric)

- This time five years ago [E6-5(R4)], Microsoft [E6-5(R2)] ***bought*** [E6-5] Nokia [E6-5(R3)] 's mobile business [E6-5(R8)] for a cool $7.6 billion [E6-5(R21)].
- Mexico's new president [E6-5(R2)] is ***selling*** [E6-5] his version of Air Force One [E6-5(R8)] and flying commercial instead for $218.7 million [E6-5(R21)].

**Event subtype (E):** Bankruptcy[6-6]

**Argument roles (R):** Participant[1] (Person, Org), Time[4] (Time), Place[5] (GP, Location)

- Wedding dress retailer David's Bridal [E6-6(R1)] ***filed for bankruptcy*** [E6-6] last month [E6-6(R4)], with the hope that restructuring would help reduce its debt.

Brookstone [E6-6(R1)] ***filed for bankruptcy*** [E6-6] in August [E6-6(R4)] and cited declining traffic in malls as one of the reasons for its demise.

**Event subtype (E):** Produce[6-7], Release[6-8]

**Argument roles (R):** Source[2] (GP, Person, Org), Target[3] (Facility), Time[4] (Time), Place[5] (GP, Location), Number of Targets[19] (Numeric)

- It is reported that China [E6-7(R2)] has succeeded in genetically modifying cows to ***produce*** [E6-7] "human" milk [E6-7(R3)].
- Apple [E6-8(R2)] Just ***Released***[E6-8] its iOS 12.1.3 Developer Beta 2 [E6-8(R3)].

**Event subtype (E):** Pricing[6-9]

**Argument roles (R):** Participant[1] (Org, Facility), Price[21] (Numeric) Time[4] (Time), Place[5] (GP, Location)

- The iPhone 11 Pro [E6-9(R1)] will ***retail*** [E6-9] for $999 [E6-9(R21)], and the 11 Pro Max [E6-9(R1)] for $1,099[E6-9(R21)].

**Event subtype (E):** Price rise[6-10], Price drop[6-11]

**Argument roles (R):** Participant[1] (Org, Facility), Time[4] (Time), Place[5] (GP, Location), Price[21] (Numeric), Fluctuation[10] (Numeric)

- As of 10:39 a.m. EDT on Monday[E6-10(R4)], WTI Crude[E6-10(R1)] was ***up***[E6-10] 0.32 percent[E6-10(R10)] at $63.30[E6-10(R21)] and Brent Crude[E6-10(R1)] was trading ***up***[E6-10] 0.19 percent[E6-10(R10)] at $66.88[E6-10(R21)].
- The price of a bitcoin[E6-11(R1)] hit a low of $52,810.06[E6-11(R21)] late Saturday[E6-11(R4)] after it[E6-11(R1)] ***tumbled***[E6-11] more than $7,000[E6-11(R10)] in a single hour, before the losses eased.

**Event subtype (E):** Production rise[6-12], Production drop[6-13]

**Argument roles (R):** Participant[1] (Facility), Time[4] (Time), Place[5] (GP, Location), Fluctuation[10] (Numeric), Number of Participants[13] (Numeric)

- The USDA forecasts that global pork[E6-12(R1)] production could ***rise***[E6-12] by 5%[E6-12(R10)] year-on-year in 2021[E6-12(R4)] to 101.5 million tonnes[E6-12(R13)], AHDB's Hannah Clarke reported.
- According to the International Organization of Motor Vehicle Manufacturers, global automotive[E6-13(R1)] production ***fell***[E6-13] by 16%[E6-13(R10)] last year[E6-13(R4)].
- Overall production of the three main cereals - wheat[E6-13(R1)], oats[E6-13(R1)] and barley[E6-13(R1)] - ***decreased***[E6-13] to 2,013,000t[E6-13(R13)] in 2020[E6-13(R4)] representing a ***decline***[E6-13] of 16%[E6-13(R10)], according to CSO figures on Area.

**Event subtype (E):** Establishment[6-14]

**Argument roles (R):** Source[2] (GP, Person, Org), Target[3] (Org), Time[4] (Time), Place[5] (GP, Location)

- Google[E6-14(R3)] was ***founded***[E6-14] in September 1998[E6-14(R4)] by Larry Page[E6-14(R2)] and Sergey Brin[E6-14(R2)] while they were Ph.D. students at Stanford University.

**Event subtype (E):** Investment[6-15], Initial Public Offering (IPO)[6-16], Capital Increase[6-17]

**Argument roles (R):** Source[2] (GP, Person, Org), Target[3] (Person, Org, Facility), Time[4] (Time), Place[5] (GP, Location), Price[21] (Numeric)

- Two-year-old Indian startup Glance[E6-15(R3)] has ***raised***[E6-15] $145 million[E6-15(R21)] in a new financing round from Google[E6-15(R2)] and existing investor Mithril Partners[E6-15(R2)].
- Trifork[E6-16(R2), E6-8(R2)] ***publishes***[E6-8] Offering Circular[E6-8(R8)] and offer price[E6-8(R8)] for its intended ***Initial Public Offering***[E6-16] on Nasdaq Copenhagen[E6-16(R3)].
- Digital Brands Group, Inc.[E6-16(R2)] Announces Pricing of $10.0 Million[E6-16(R21)] ***Initial Public Offering***[E6-16] and Nasdaq Listing[E6-16(R3)].
- Air France-KLM[E6-17(R3)] announces the success of its ***capital increase***[E6-17] for an amount of €1.036 billion[E6-17(R21)] after exercise in full of the increase option.
- GOL Linhas Aéreas Inteligentes S.A. Brazil's largest domestic airline[E6-17(R3)], today announces it is initiating a ***capital increase***[E6-17] of up to approximately R$512 million[E6-17(R21)] led by its controlling shareholders[E6-17(R2)].

## 7) Event type: Politics

**Event subtype (E):** Travel[7-1]

**Argument roles (R):** Source[2] (Org, Person), Target[3] (GP, Location), Time[4] (Time), Place[5] (GP, Location), Vehicle[7] (Facility)

- ARAS ON A SEAT President of Ireland Michael D Higgins[E7-1(R2)] ***travels***[E7-1] to Gran Canaria[E7-1(R3)].
- U.S. President Donald Trump[E7-1(R2)] ***arrives***[E7-1] in Saudi Arabia[E7-1(R3)], his first of three scheduled foreign trips.

**Event subtype (E):** Cooperation[7-2], End cooperation[7-3], Aid[7-4], Import[7-5], Export[7-6], settlement[7-7]

**Argument roles (R):** Source[2] (GP, Person, Org), Target[3] (GP, Person, Org), Time[4] (Time), Place[5] (GP, Location), Artifact[8] (Animal, Facility), Price[21] (Numeric)

- Saudi Arabia[E7-2(R2)], the United Arab Emirates[E7-2(R2)], Egypt[E7-2(R2)], Bahrain[E7-2(R2)], Yemen[E7-2(R2)] and the Maldives[E7-2(R2)] say they[E7-2(R2)] are ***severing diplomatic relations***[E7-2] with Qatar[E7-2(R3)].
- Ukraine[E7-3(R2)] ***cut off cooperation***[E7-3] with Mueller[E7-3(R3)] to Get Missiles from Trump.
- Japan[E7-4(R2)] has decided to ***help***[E7-4] the fifth tranche of assistance worth $4 million[E7-4(R21)] to Pakistan[E7-4(R3)] to support its efforts to combat Covid-19.
- South Korea[E7-5(R2)] plans to ***import***[E7-5] arms[E7-5(R8)] worth about 3.5 trillion won[E7-5(R21)] this year[E7-5(R4)] to enhance its defense capabilities.
- Malaysia[E7-5(R3)] bans the ***import***[E7-5] of horses[E7-5(R8)] from Australia[E7-5(R2), E3-2(R5)] following an ***outbreak***[E3-2] of Hendra virus[E3-2(R2)].
- In 2012[E7-6(R4)], Iran[E7-6(R2)], which ***exported***[E7-6] around 1.5 million barrels of crude oil[E7-6(R8)] a day[E7-6(R4)], was the second-largest exporter among the Organization of Petroleum Exporting Countries.

- Trump[E7-7(R2)] ***signs*** [E7-7] an arms deal[E7-7(R8)] with King Salman [E7-7(R3)] worth more than US$100 billion[E7-7(R21)].

**Event subtype (E):** Breach of settlement[7-8], End settlement [7-9]

**Argument roles (R):** Source[2] (GP, Person, Org), Target[3] (GP, Person, Org), Time[4] (Time), Place[5] (GP, Location), Artifact[8] (Facility)

- Russia [E7-8(R2)] tested a ground-launched cruise missile, ***breaking*** [E7-8] the Intermediate-Range Nuclear Forces Treaty [E7-8(R8)] ***signed*** [E7-7] in 1987 during the Cold War [E7-7(R4)], the US said.
- The US[E7-9(R2)] has become the first nation in the world to formally ***withdraw***[E7-9] from the Paris climate agreement **[E7-9(R8)]**.

**Event subtype (E):** Meeting[7-10], Military Exercise[7-11]

**Argument roles (R):** Participant[1] (GP, Person, Org), Time[4] (Time), Place[5] (GP, Location), Number of Participants[13] (Numeric)

- Donald Trump's secret[E7-1(R2), E7-10(R1)] ***flight***[E7-1] to Iraq[E7-1(R3), E7-10(R5)] to ***visit***[E7-10] troops[E7-10(R1)] tracked online by aircraft spotters.
- U.S. President Donald Trump[E7-10(R1)] ***meets***[E7-10] with North Korean leader Kim Jong Un[E7-10(R1)] on Sentosa Island, in Singapore[E7-10(R5)].
- A large-scale Russian[E7-11(R1)] ***military exercise***[E7-11] is coming to the Arctic[E7-11(R5)].

**Event subtype (E)**: Sanction[7-12], Ban[7-13]

**Argument roles (R):** Source[2] (GP, Person, Org), Target[3] (GP, Person, Org, Facility), Time[4] (Time), Place[5] (GP, Location), Duration[12] (Time)

- New ***sanctions*** [E7-12] against North Korea [E7-12(R3)], issued by the Department of Treasury [E7-12(R2)] on Dec. 10[E7-12(R4)].
- The U.S. Department of the Treasury[E7-12(R2)] ***sanctioned*** [E7-12] three high ranking North Korean officials [E7-12(R3)] on Monday [E7-12(R4)].
- Google [E7-13(R2)] announces it [E7-13(R2)] will ***ban*** [E7-13] targeted ads that use voter data [E7-13(R3)], effective within a week [E7-13(R4)].
- English Soccer [E7-13(R2)] Will ***Boycott*** [E7-13] Social Media [E7-13(R3)] to Protest Online Abuse.
- China[E7-13(R2)] ***blocks***[E7-13] Microsoft's Bing search engine[E7-13(R3)], according to a new report.

**Event subtype (E)**: Threat[7-14], Extradite[7-15], Exile[7-16], Apologize[7-17], Deport[7-18], Interference[7-19], Troops withdrawal[7-20], Criticism[7-21], Condemn[7-22], Dissolution[7-23], Refuge[7-24], Conflict[7-25], Conquering[7-26], Occupy[7-27]

**Argument roles (R):** Source[2] (GP, Person, Org), Target[3] (GP, Person, Org, Facility), Time[4] (Time), Place[5] (GP, Location)

- North Korea [E7-14(R3)] says it won't denuclearize unless US [E7-14(R2)] removes ***threat*** [E7-14].
- North Korea[E7-14(R2)] launched its first-ever intercontinental ballistic missile and ***threatened***[E7-14] to send more missiles.
- Czech authorities [E7-15(R2)] ***extradite*** [E7-15] from the United States [E7-15(R5)] Kevin Dahlgren [E7-15(R3), E4-9(R2)], who is suspected of ***killing*** [E4-9] four [E4-9(R19)] Czech family member [E4-9(R3)].
- Turkey issues an arrest warrant for Fethullah Gülen [E7-16 (R3)], who currently lives in ***self-imposed exile*** [E7-16] in the United States [E7-16(R5)].
- United States President Barack Obama[E7-10(R1), E7-17(R2)] ***telephoned***[7-10] MSF International President Joanne Liu [E7-10(R1), E7-17(R3)] to ***apologize*** [E7-17] for the U.S. [E4-1(R2)] ***bombing*** [E4-1] of the hospital [E4-1(R3)] in Afghanistan [E4-1(R5)].
- Myanmar [E7-18(R2)] announces plans to ***deport*** [E7-18] recently rescued migrants [E7-18(R3)] to Bangladesh [E7-18(R5)].
- FIFA [E7-13(R2)] ***suspends***[E7-13] the Nigeria Football Federation [E7-13(R3), E7-19(R3)] due to government [E7-19(R2)] ***interference*** [E7-19].
- Biden [E7-20(R2)] announces U.S. troops [E7-20(R3)] to ***leave*** [E7-20] Afghanistan [E7-20(R5)] by Sept. 11 [E7-20(R4)].
- Zarif [E7-21(R2)] ***criticised*** [E7-21] France [E7-21(R3), E7-8(R2)], Germany [E7-21(R3), E7-8(R2)] and Britain [E7-21(R3), E7-8(R2)] for ***failing to enforce the agreement*** [E7-8] since 2018 [E7-8(R4)], when U.S. President Donald Trump [E7-9(R2), E7-12(R2)] ***abandoned*** [E7-9] the deal [E7-9(R8)] and restored ***harsh economic sanctions*** [E7-12] on Iran [E7-9(R3), E7-12(R3)].
- Chicago activists[E7-22(R2)] ***condemn***[E7-22] Lightfoot's National Guard request [E7-22(R3)], demand charges be dropped against protesters [E5-2(R3), E8-2(R2)] ***arrested*** [E5-2] in Logan Square [E5-2(R5), E8-2(R5)] ***march*** [E8-2].
- The military leaders [E7-23(R2)] announce the ***dissolution*** [E7-23] of the Thai Senate [E7-23(R3)].

- Blind Chinese human rights activist Chen Guangcheng [E4-4(R2), E7-24(R2)] – who ***fled*** [E4-4] from house arrest [E4-4(R5)] – reportedly ***takes refuge*** [E7-24] in the U.S. embassy [E7-24(R3)] in Beijing [E7-24(R5)].
- ***The ongoing conflict*** [E7-25] between India [E7-25(R2), E7-28(R2)] and Pakistan [E7-25(R2), E7-28(R2)] had brought the two nuclear-armed nations ***close to war***[E7-28] on more than one occasion.
- The Achaemenid monarch Cambyses [E7-26(R2)] ***conquered***[E7-26] Egypt [E7-26(R2)] in the year 525 BC[E7-26(R4)].
- Groups of indigenous peoples [E7-27(R2)] in Brazil [E7-27(R5)] **occupy** [E7-27] government buildings [E7-27(R3)].

**Event subtype (E):** War[7-28], End-war[7-29]

**Argument roles (R):** Source[2] (GP, Person, Org), Target[3] (GP, Person, Org), Time[4] (Time), Place[5] (GP, Location), Number of Injuries[14] (Numeric), Number of Deaths[15] (Numeric), Number of Missing Entities[16] (Numeric), Number of Destructions[17] (Numeric)

- The Iran[E7-28(R2)] – Iraq [E7-28(R2)] ***War***[E7-28] began on 22 September 1980[E7-28(R4)].
- On March 26, 1979 [E7-7(R4), E7-29(R4)], Israel [E7-7(R2), E7-29(R2)] and Egypt [E7-7(R2), E7-29(R2)] ***signed a peace treaty*** [E7-7] formally ***ending the state of war*** [E7-29] that had existed between the two countries [E7-29(R2)] for 30 years [E7-29(R4)].

**Event subtype (E):** Suppress [7-30]

**Argument roles (R):** Source[2] (GP, Person, Org), Target[3] (Person, Org), Instrument[6] (Facility), Time[4] (Time), Place[5] (GP, Location), Number of Injuries[14] (Numeric), Number of Deaths[15] (Numeric), Number of Sources[18] (Numeric), Number of Targets[19] (Numeric)

Myanmar Forces [E7-30(R2)] ***Suppress*** [E7-30] ***Protests*** [E8-1], ***Kill***[E1-1] More Than 100 [E1-1(R13), E7-30(R15), E8-1(R15)] People [E1-1(R1), E7-30(R3), E8-1(R2)].

**Event subtype (E):** Impeachment [7-31]

**Argument roles (R):** Source[2] (Person, Org), Target[3] (Person), Time[4] (Time), Place[5] (GP, Location)

- President of Peru Martín Vizcarra[E7-31(R3)] is ***impeached***[E7-31] by the Congress of Peru [E7-31(R2)].

## 8) Event type: Social

**Event subtype (E):** Protest[8-1], Coups[8-2]

**Argument roles (R):** Source[2] (GP, Person, Org), Target[3] (GP, Person, Org), Time[4] (Time), Place[5] (GP, Location), Number of Sources[18] (Numeric), Number of Targets[19] (Numeric), Number of Injuries[14] (Numeric), Number of Deaths[15] (Numeric), Number of Missing Entities[16] (Numeric), Number of Destructions[17] (Numeric)

- Hundreds of thousands[E8-1(R18)] ***demonstrate***[E8-1] throughout Venezuela[E8-1(R5)] against PresidentNicolás Maduro[E8-1(R3)].
- Turkish police[E4-1(R2)] ***fire***[E4-1] tear gas[E4-1(R6)] and rubber bullets[E4-1(R6)] at demonstrators [E4-1(R3), E8-1(R2)] ***marching***[E8-1] to Taksim Square[E4-1(R5), E8-1(R5)].
- Dozens[E1-1(R13), E8-2(R15)] ***killed*** [E1-1] in foiled Ethiopia[E1-1(R5), E8-2(R5)] ***coup attempt*** [E8-2], authorities say.

**Event subtype (E):** Ceremony[8-3], Revolution[8-4]

**Argument roles (R):** Participant[1] (GP, Person, Org), Time[4] (Time), Place[5] (GP, Location), Number of Participant[13] (Numeric)

- New York City [E8-3(R5)] begins ***the 50th Annual World Pride festival*** [E8-3], expecting to draw four million [E8-3(R13)] people [E8-3(R1)] over six days [E8-3(R4)].
- ***A ceremony*** [E8-3] in March[E8-3(R4)] marking the closure of the last makeshift hospital in Wuhan, China[E8-3(R5)], where the new coronavirus first emerged.
- In December 2010 [E1-1(R4), E8-1(R4), E8-4(R4)], ***protests*** [E8-1] in Tunisia [E1-1(R5), E8-1(R5), E8-4(R5)] sparked by the ***death*** [E1-1] of Mohamed Bouazizi [E1-1(R1)] turned into a ***revolution*** [E8-4].

## 9) Event type: Cyberspace

**Event subtype (E):** CyberAttack[9-1], Information Disclosure[9-2]

**Argument roles (R):** Source[2] (GP, Org, Person, Facility), Target[3] (Person, Org, Facility), Instrument[6] (Facility), Time[4] (Time), Place[5] (GP, Location), Number of Targets[19] (Numeric)

- ***A large scale international cyber-attack*** [E9-1] using the WannaCry ransomware package [E9-1(R6), E4-6(R6)] ***disrupts*** [E4-6] computer and telephone systems [E9-1(R3), E4-6(R3)] across 99 countries [E9-1(R5), E4-6(R5)].
- The Italian oil-services company Saipem[E9-1(R3)] is still assessing the scope and impact of ***a cyberattack***[E9-1] that targeted its servers[E9-1(R3)] in the Middle East [E9-1(R5)], according to the head of digital and innovation.
- Hackers[E9-(R2)] ***compromised*** [E9-2] 38 million [E9-2(R19)] Chinese users data [E9-2(R3)].

## 10) Event type: Election

**Event subtype (E):** Holding elections[10-1]

**Argument roles (R):** Participant[1] (GP, Person, Org), Time[4] (Time), Place[5] (GP, Location), Number of Participants[13] (Numeric), Occupation[9] (Occupation)

- Voters[E10-1(R1)] in Haiti [E10-1(R5)] go to the polls for the second round of voting in ***the presidential election*** [E10-1].

**Event subtype (E):** Election results[10-2]

**Argument roles (R):** Participant[1] (GP, Person, Org), Time[4] (Time), Place[5] (GP, Location), Number of Participants[13] (Numeric), Occupation[9] (Occupation)

- Incumbent President Hassan Rouhani [E10-2(R1)] is ***re-elected*** [E10-2] with 23.5 million [E10-2(R13)] votes.
- Macky Sall , former Prime Minister under Abdoulaye Wade 's administration [E10-2(R1)], is ***elected*** [E10-2] President [E10-2(R9)] of Senegal [E10-2(R5)].

**Event subtype (E):** Election Candidate[10-3]

**Argument roles (R):** Source[2] (GP, Org, Person), Target[3] (Facility), Time[4] (Time), Place[5] (GP, Location)

- The 2020 presidential campaign of Joe Biden [E6-14(R3)] ***began*** [E6-14] on April 25, 2019 [E6-14(R4), E6-8(R4), E10-3(R4)], when Biden [E10-3(R2), E6-8(R2)] ***released*** [E6-8] a video [E6-8(R8)] ***announcing his candidacy*** [E10-3] in the 2020 Democratic party presidential primaries [E10-3(R3)].

## 11) Event type: Transport accident

**Event subtype (E):** Traffic Collision [11-1], Rail accidents[11-2], Aviation accidents[11-3], Marine accidents[11-4]

**Argument roles (R):** Participant[1] (Person, Animal), Time[4] (Time), Place[5] (GP, Location), Vehicle[7] (Facility), Number of Participants[13] (Numeric), Number of Injuries[14] (Numeric), Number of Deaths[15] (Numeric), Number of Missing Entities[16] (Numeric)

- A car [E11-1(R7)] ***crash*** [E11-1] in Billerica, Massachusetts [E11-1(R5), E1-1(R5), E1-2(R5)], ***kills*** [E1-1] at least three [E11-1(R15), E1-1(R13)] and ***injures*** [E1-2] nine [E11-1 (R14), E1-2(R13)].
- ***A collision*** [E11-2] of two trains [E11-2(R7)] near Aichach in Germany [E11-2(R5), E1-1(R5), E1-2(R5)] leaves two [E11-2(R15), E1-1(R13)] people [E11-2(R1), E1-1(R1), E1-2(R1)] ***dead*** [E1-1] and at least 14 [E11-2(R14), E1-2(R13)] ***injured*** [E1-2].
- An Ethiopian Airlines  Boeing 737 MAX 8 from Addis Ababa , Ethiopia [E11-3(R7)], with 149 [E11-3(R13)] passengers [E11-3(R1)] and eight [E11-3(R13)] crew [E11-3(R1)] members onboard ***crashes*** [E11-3] en route to Nairobi , Kenya [E11-3(R5), E1-1(R5)], ***killing*** [E1-1] all 157 [E11-3(R15),E1-1(R13)] persons [E11-3(R1),E1-1(R1)] on board.
- A boat [E11-4(R7)] ***capsized*** [E11-4] on Monday [E11-4(R4), E1-1(R4)] in a lake near Egypt's Mediterranean city of Alexandria [E11-4(R5), E1-1(R5)], leaving at least five [E11-4(R15), E1-1(R13)] people [E11-4(R1), E1-1(R1)] ***dead*** [E1-1].

**Event subtype (E):** Collapse[11-5], Hazardous material spill[11-6], Fire[11-7]

**Argument roles (R):** Participant[1] (Facility, Person, Animal), Time[4] (Time), Place[5] (GP, Location), Number of Participants[13] (Numeric), Number of Injuries[14] (Numeric), Number of Deaths[15] (Numeric), Number of Missing Entities[16] (Numeric), Number of Destructions[17] (Numeric)

- Dozens [E1-1(R13), E11-5(R15), E11-7(R15)] of firefighters [E1-1(R1)] were ***killed*** [E1-1] when a high-rise building [E11-5(R1), E11-7(R1)] ***collapsed*** [E11-5] after ***a blaze*** [E11-7] in Iran's capital [E1-1(R5), E11-5(R5), E11-7(R5)], state-run Press TV reported Thursday [E1-1(R4), E11-5(R4), E11-7(R4)].
- Seven [E1-1(R13), E11-1(R15), E11-6(R15), E11-7(R15)] people [E1-1(R1), E11-1(R1)] are ***killed*** [E1-1] after ***a crash*** [E11-1] and diesel fuel [E11-6(R1)] ***spill*** [E11-6] sparked ***a massive fire*** [E11-7] on Interstate 75 in Gainesville, Florida , United States [E1-1(R5), E11-1(R5), E11-6(R5), E11-7(R5)].
- The UN says approximately 500 [E4-1(R18)] gunmen [E4-1(R2)] ***attacked*** [E4-1] a Masalit community [E4-1(R3)] in West Darfur [E4-1(R5), E11-7(R5), E1-1(R5)], ***burning down*** [E11-7] houses [E11-7(R1), E4-1(R3)], and ***killing*** [E1-1] more than 60 [E1-1(R13), E4-1(R15)] people [E1-1(R1), E4-1(R3)].

## 12) Event type: Science

**Event subtype (E):** Invention[12-1]

**Argument roles (R):** Source[2] (GP, Person, Org), Time[4] (Time), Place[5] (GP, Location), Artifact[8] (Facility)

- The Association for Computing Machinery awards the 2015 A.M. Turing Award to Whitfield Diffie [E12-1(R2)] and Martin Hellman [E12-1(R2)] for ***the invention*** [E12-1] of public-key cryptography[E12-1(R8)] and digital signatures[E12-1(R8)] which revolutionized computer security.

**Event subtype (E):** Discovery[12-2]

**Argument roles (R):** Source[2] (GP, Person, Org), Target[3] (Facility), Time[4] (Time), Place[5] (GP, Location), Number of Targets[19] (Numeric)

- The Egyptian Ministry of Antiquities [E12-2(R2)] announces ***the discovery*** [E12-2] of eight [E12-2(R19)] mummies [E12-2(R3)], 10 [E12-2(R19)] colorful sarcophagi [E12-2(R3)], and numerous [E12-2(R19)] figurines [E12-2(R3)].
- NASA and the European Southern Observatory [E12-2(R2)] announce ***the discovery*** [E12-2] of four[E12-2(R19)] new Earth-like planets [E12-2(R3)] in the Goldilocks zone of the star TRAPPIST-1 [E12-2(R5)].

*Table 9. Representing words of event subtypes that are used for testing the generality of COfEE on Wikipedia events*

| Type | Subtype | Subtype display in Figures 7 and 11 | Event Words | Count |
|---|---|---|---|---|
| **Accident** | Collapse | Collapse | collapse, collapses, collapsed, collapsing | 459 |
| **Accident** | Fire | Fire | blaze, fire, wildfire, wildfires, forest fires, bushfire, bushfires, burning, burn, burns, burnt, burned, burnings, building fire, structure fire, construction fire | 2,560 |
| **Accident** | Hazardous Material Spill | Hazardous Material Spill | spill, spills, spilled, leak of radioactive, radioactive leak, radioactive leakage, radioactive leaks | 248 |
| **Accident** | Rail Accidents, Marine Accidents, Aviation Accidents, Traffic Collision | Transport Accident (Group) | rail accident, rail accidents, accident, accidents, crash, crashes, crashed, crashes, collision, collisions, wreck, wrecks | 1,540 |
| **Business** | Bankruptcy | Bankruptcy | bankruptcy, bankrupt, bankrupted, bankrupts, bankrupting | 129 |
| **Business** | Capital Increase | Capital Appreciation | capital increase, capital increment | 1 |
| **Business** | End Position | End Position | dismissal, dismissals, resign, resigns, resigning, resigned, resignation, dismiss, expulsion, expulsions, ouster, ousters, retire, retires, retired, retirement, retirements, expel, expels, expelling, expelled, fired | 1,829 |
| **Business** | Establishment | Establishment | establishment, establishments, establish, established, establishing, creation | 258 |
| **Business** | Investment | Investment | investment, investments, capital, capitals, funding, fundings | 1,184 |
| **Business** | IPO | IPO | initial public offering, ipo, ico, initial coin offering | |
| **Business** | Money Transfer | Money Transfer | transfer money, money transfer, payment, payments, repayment, repayments, pay, pays, paid, paying | 537 |
| **Business** | Ownership Transfer | Ownership Transfer | ownership transfer, ownership transfers, ownerships transfer, ownership unbundle, ownership unbundling, ownership unbundlings, sell, sells, sold, selling, sale, sales, auction, auctioned, auctions, auctioning, buy, buys, bought, buying, purchase, purchased, purchases | 2,040 |
| **Business** | Price Drop | Price Drop | price drop, prices drop, price drops, price dropping, price decrease, price decreases, price decreased, prices decrease, price decreasing, price decrement, prices decrement, price decrements | 5 |
| **Business** | Price Rise | Price Rise | price rise, prices rise, price hike, prices hike, price hikes, inflation, inflated, inflate, inflating, price increase, price increases, price increased, prices increase, price increasing, price boost, boost price | 51 |
| **Business** | Pricing | Pricing | pricing, price, prices, priced, tariff, tariffs | 259 |
| **Business** | Produce, Production rise | Production (Group) | produce, produces, produced, productions, development, developments, develop, develops, developed, production rise, productions rise, production rises, production rising, production increase, production increasing, production increased, productions increase, productions increasing, raise production, raise productions, raises production, production raise, productions raise, boost production, production boost | 371 |
| **Business** | Production Drop | Production Drop | production drop, productions drop, production drops, productions dropped, production dropped, production decrement, productions decrement, production decrements, production decrease, production decreasing, productions decrease, productions descreased, production decreased, stop production, production stop, cut production, production cut, suspend production, production suspend, production suspension, | 41 |
| | | | end production, production end, shut production, production shut, halt production, production halt, cease production, production cease | |
| **Business** | Recruitment | Recruitment | recruitment, recruit, recruiting, recruited, recruiter | 57 |
| **Business** | Release | Release | release, releases, releasing, released, launch, launches, launching, launched, unveil | 2,056 |
| **Business** | Start Position | Start Position | appointing, appoint, appoints, appointed, appointment, appointments | 338 |
| **Crime** | Attack | Attack | attack, attacks, attacking, attacked, fight, fights, fighting, fought, stab, stabs, stabbing, stabbed, assault, assaults, assaulting, assaulted, combat, combated, combating, combatted, combats, combatting, shoot, shootings, shoots, shooting, bomb, bombing, bombs, bombed, airstrike, airstrikes, battle, battles, counterattack, shooting spree, attempted murder, shot, shots, bomber, missile strikes, missile, missiles, rocket | 9,019 |
| **Crime** | Copyright Violation | Copyright Violation | copyright violation, copyright violations, copyright violate, copyright violated, copyright infringement, copyright infringements, copyright infringements, copyright infringer, copyright infringers, copyrights violation, copyrights violations, copyrights violate, copyrights violated, copyrights infringement, copyrights infringements, copyrights infringements, copyrights infringer, copyrights infringers | 11 |
| **Crime** | Destruction | Destruction | destruction, destructions | 126 |
| **Crime** | Economic Corruption | Economic Corruption | economic corruption, economic corruptions, abusing, abuse, abuses, abused, bribing, bribes, bribery, overcharging, overchargings, money laundering, fraud, frauds, corruption, scandal, scandalism, scandals, laundering, tax evasion, fraudulent, scam, scams, embezzlement, embezzlement, hoarding, hoard | 1,479 |
| **Crime** | Escape | Escape | escape, escapes, escaping, escaped, fleeing, flee, flees, fled | 447 |
| **Crime** | Espionage | Espionage | espionage, spying, spy, spies, counterspies, counterintelligence, cyberespionage, cyberspies, spycraft, covert ops, eavesdropping, counterspy, spymaster, spymasters | 194 |
| **Crime** | Explosion | Explosion | explosion, explosions, explode, exploded, blast, blasts, bomb detonated | 1,109 |
| **Crime** | Homicide | Homicide | homicide, homicides, murder, murdered, murders, behead, beheaded, beheading, assassinate, assassinated, assassination, assassinations, genocide, genocides | 1,127 |
| **Crime** | Hostage Crisis | Hostage Crisis | hostage crisis, hostage crises, hostage taking, hostage takings, hostage, hostages | 212 |
| **Crime** | Human-rights Violation | Humanrights Violation | human right violation, human right violations, human right violate, human right violated, human right violator, humanright violation, humanright violations, humanright violate, humanright violated, humanright violator, humanrights violation, humanrights violations, humanrights violate, humanrights violated, humanrights violator, human rights violation, human rights violations, human rights violate, human rights violated, human rights violator, human rights abuse, human right abuse, humanright abuse, humanrights abuse | 61 |
| **Crime** | Kidnapping | Kidnapping | kidnapping, kidnappings, kidnap, kidnapped, kidnapping, kidnaps, kidnappers, abducted, abduction, abductions, captive, captives | 364 |
| **Crime** | Privacy Violation | Privacy Violation | privacy, privacy violation, privacy violations, violation of privacy, violations of privacy | 43 |
| **Crime** | Robbery | Robbery | robbery, robberies, carjacking, carjackings, burglary, burglaries, robber, robbers | 70 |
| **Crime** | Sex Assault | Sex Assault | sex assault, sex assaults, rape, rapes, raped | 219 |
| **Crime** | Smuggling | Smuggling | smuggling, smuggle, smuggles, smuggled, smugglers, trafficking, traffickers | 200 |
| **Crime** | Torture | Torture | torture, torturing, tortured, tortures, torturer, torturers | 129 |
| **Cyberspace** | CyberAttack, Information Disclosure | Cyberspace (Group) | cyberattack, cyber attack, cyber attacks, cyberattacks, hack, hacking, hacked, hacker, hackers, privacy, data breach, data breachs, information leakage, leak information, disclosure of information, information disclosure | 372 |
| **Election** | Election Candidacy, Election Results, Holding Election | Election (Group) | election candidacy, election campaign, candidacy, candidate, candidates, nominate, nominating, nominates, nominated, nomination, nominations, nominee, nominees, election, elections, elect, elected, poll, polls, referendum, referendums, voter, voters | 4,233 |
| **Environment** | Emergency Evacuation | Emergency Evacuation | evacuation, evacuations, evacuate, evacuates, evacuating, evacuated | 545 |
| **Environment** | Epidemics | Epidemics | epidemic, epidemics, pandemic, pandemics, covid, covid19, covid-19, virus, viruses, disease, diseases, outbreak, outbreaks | 3,692 |
| **Environment** | Extinction | Extinction | extinction, extinctions, extinct, critically endangered | 32 |
| **Environment** | Hunting | Hunting | hunting, huntings, hunters, hunt animal, hunt animals, animal hunter, animal hunters | 23 |
| **Environment** | Pollution | Pollution | pollution, pollutions, polluted, polluting, pollute, emissions, smog, smogs | 127 |
| **Environment** | Quarantine | Quarantine | quarantine, quarantined, quarantines, quarantining, biosecurity protocols, strict_biosecurity, lockdown, lockdowns, stringent biosecurity, infected zone | 374 |
| **Environment** | Resource Shortage | Resource Shortage | outage, outages, outtage, outtages, blackout, blackouts, powercut, powercuts, brownout, brownouts, internet disruption, internet disruptions, internet disrupted, internet interruptions, internet interruption, internet shutdown, internet shutdowns, internet outage, internet outages, internet restriction, internet restrictions, famine, hungry, hunger, starvation, starvations, malnutrition, malnutritions | 200 |
| **Justice** | Arrest | Arrest | arrest, arrests, arresting, arrested, capture, captures, capturing, apprehended, detained, seize | 2,823 |
| **Justice** | Complaint | Complaint | complaint, complaints, lawsuit, accuse, accuses, accused, accusation, accusations | 998 |
| **Justice** | Execution | Execution | execution, executions, execute, executed, executes, lethal injection | 225 |
| **Justice** | Fining | Fining | fine, fines, fined, fining | 268 |
| **Justice** | Imprisonment | Imprisonment | imprisonment, jailed, jail, prison, prisoned, imprisoned | 1,418 |
| **Justice** | Pardoning | Pardoning | pardon, pardons, pardoned, clemency, clemencies | 64 |
| **Justice** | Prisoner Release | Prisoner Release | prisoner release, prisoners release, prisoner released, prisoner releasing, prisoners releasing, prisoners released, release prisoner, release prisoners, released prisoner, released prisoners, releasing prisoner, releasing prisoners, prisoners freed, free prisoner, free prisoners, prisoners free | 10 |
| **Justice** | Prosecution | Prosecution | prosecution, prosecutors, prosecutor, prosecuting, prosecute, prosecutes, prosecuted | 290 |
| **Justice** | Surrender | Surrender | surrender, surrenders, surrendered, relinquish, relinquished, capitulate, capitulated, capitulating | 69 |
| **Justice** | Trial | Trial | trial, retrial, trials, retrials | 649 |
| **Life** | Birth | Birth | birth, births, birthed, birthing, born, newborn, newborns, childbirth, childbirths | 237 |
| **Life** | Death | Death | death, deaths, die, dies, dying, died, dead, kill, killed, killing, kills, funeral, funerals, slay, slays, slayed, slain, decapitate, decapitates, decapitated, corpse, corpses, massacre, massacres, massacred | 14,041 |
| **Life** | Divorce | Divorce | divorce, divorced, divorces, divorcing | 16 |
| **Life** | Drowning | Drowning | drowning, drown, drowns, drowned | 69 |
| **Life** | Hospitalization | Hospitalization | hospitalization, hospitalizations, hospitalize, hospitalized, hospitalizes, hospitalizing, surgery | 199 |
| **Life** | Immigration | Immigration | emigration, emigrate, emigrated, emigrates, emigrating, migrate, migrating, migrates, migrated, immigrate, immigrating, immigrated, immigrates, immigration | 122 |
| **Life** | Injury | Injury | injury, injuries, poison, poisons, poisoning, poisoned, wounded, mutilate, mutilates, mutilated, mutilation, mutilating, injure, injures, injuring, injured, hurt, hurts, hurting, harm, harms, harming, illness, illnesses, sickness | 4,219 |
| **Life** | Marriage | Marriage | marriage, marriages, marry, marrying, marries, married | 215 |
| **Life** | Missing | Missing | missing, miss, misses, missed, lost, disappear, disappeared, disappears, disappearance | 2,645 |
| **Life** | Suicide | Suicide | suicide, suicided, suicides, suiciding | 780 |
| **Life** | Survival | Survival | survival, survivals, rescue, survive, surviving, survives, rescue, rescues, rescuing, rescued, rescuers | 561 |
| **Natural Disasters** | Avalanche | Avalanche | avalanche, avalanches | 61 |
| **Natural Disasters** | Bad Weather | Bad Weather | bad weather, haze, hazy, thundersnow, thundersnows, heavy rain, heavy rains, snow, snowy, snows, snowing, snowed, freezing, freeze, freezes, frozen, heatwave, heatwaves, lightning, lightnings, rain, rains, tropical depression, wind, winds, windy, mist, misty, heavy rainfall, torrential rain, snowfall, heavy snowfall, cold weather, heatwave, heat waves, heatwaves, hot weather, fog, foggy | 2,911 |
| **Natural Disasters** | Drought | Drought | drought, droughts, dry spell, dry spells, dryspell, dryspells | 45 |
| **Natural Disasters** | Earthquake | Earthquake | earthquake, earthquakes, quake, aftershock | 597 |
| **Natural Disasters** | Flood | Flood | flood, floods, flooding, flooded | 769 |
| **Natural Disasters** | Landslide | Landslide | landslide, landslides, mudslide, mudslides, landfall | 466 |
| **Natural Disasters** | Storm | Storm | storm, storms, storming, stormed, thunderstorm, thunderstorms, sandstorm, sandstorms, cyclone, cyclones, tropical storm, tropical storms, duststorm, duststorms, winter storm, winter storms, hurricane, hurricanes, tornado, tornados, tornadoes, snowstorm, snowstorms, typhoon, typhoons, blizzard | 1,077 |
| **Natural Disasters** | Tsunami | Tsunami | tsunami, tsunamis | 157 |
| **Natural Disasters** | Volcanic Eruption | Volcanic Eruption | volcanic eruption, eruption, eruptions, erupt, erupts, erupting, erupted, volcano, volcanoes, volcanic eruptions, lava, volcanic | 279 |
| **Politics** | Aid | Aid | aid, aids, aiding, defend, defends, defending, defended, defense, defenses, bailout, bailing out, bailouts, bailed out, bails out, assistance, assistances, relief, reliefs, protect, defensive, defenders | 2,384 |
| **Politics** | Apologize | Apologize | apologize, apologizes, apologizing, apologized, apology | 93 |
| **Politics** | Ban | Ban | boycott, boycotts, boycotting, boycotted, ban, bans, banned, banning, censorship, censorships, censoring, block, blocking, blocked, suspension, suspensions | 3,366 |
| **Politics** | Condemn | Condemn | condemn, condemns, condemning, condemned, denounce, denounced, denouncing | 345 |
| **Politics** | Conflict | Conflict | conflict, conflicts, disagreement, disagreements, dispute, disputes, disputed, clash, clashes, clashed, strife, strifes, hostility, hostilities, violation, feud, feuding, feuded, feuds, standoff, standoffs, fight, combat | 3,382 |
| **Politics** | Conquering | Conquering | conquering, conquer, conquered, capture, captures, capturing, recapture, recaptures, recaptured, seize, seized, seizes, seizing, regain, reclaim | 775 |
| **Politics** | Cooperation | Cooperation | cooperation, cooperations, cooperate, cooperates, co-operates, co-operated, co-operating, collaboration, alliance, alliances, trade, trades, trading, traded, partnership, collaborating, accord, coalition, coalitions, allies | 1,711 |
| **Politics** | Criticism | Criticism | criticism, criticisms, criticizm, criticizms, criticize, criticize, criticizes, criticises, criticised, criticized, criticising, criticizing | 335 |
| **Politics** | Deport | Deport | deport, deports, deported, deporting | 108 |
| **Politics** | Dissolution | Dissolution | dissolve, dissolved, dissolution, dissolving, dissolutions, disbanding, dissolves, disband, disbands | 128 |
| **Politics** | End Cooperation | End Cooperation | end cooperation, cooperation end, cooperations end, ends cooperation, finish cooperation, stop cooperation, finish cooperations, stop cooperations, suspend cooperation, suspends cooperation, suspend cooperations, suspends cooperations, cooperation suspension, cooperation suspend, cooperations suspension, cooperations suspend | 5 |
| **Politics** | End war | End war | make peace, ceasefire, peace, truce | 802 |
| **Politics** | Exile | Exile | exile, exiled, withdrawal, withdraws | 384 |
| **Politics** | Export | Export | export, exports, exported, exporting | 83 |
| **Politics** | Extradite | Extradite | extradite, extradites, extradited, extradition | 156 |
| **Politics** | Impeachment | Impeachment | impeachment, impeachments | 83 |
| **Politics** | Import | Import | import, imports, imported, importing, importation | 182 |
| **Politics** | Interference | Interference | interference, interferences, intervention, interventions | 458 |
| **Politics** | Meeting | Meeting | meeting, meet, meets, met, meetings, conference, conferences, summit, summits, visit, visits, visited, visiting, talked, talking, talkings, negotiation, negotiations, debate, debated, debates, interview, interviews, interviewed, interviewing, briefing, briefings, forum | 2,735 |
| **Politics** | Military Exercise | Military Exercise | military exercise, parade, parades | 105 |
| **Politics** | Occupy | Occupy | occupy, occupying, occupies, occupied, reoccupy | 97 |
| **Politics** | Refuge | Refuge | refuge, refuges, refuged, refuging, refugee, refugees | 303 |
| **Politics** | Sanction | Sanction | sanction, sanctions, sanctioned | 367 |
| **Politics** | Settlement, Breach of Settlement, End Settlement | Settlements (Group) | settlement, agreement, agreements, contract, contracts, contracting, contracted, sign, signs, ties, treaty, pact, pacts, deal, deals, resoloution (law) | 5,815 |
| **Politics** | Suppress | Suppress | suppress, suppresses, suppressing, suppressed, suppression | 25 |
| **Politics** | Threat | Threat | threat, threats, threaten, threatens, threatened, threatening | 679 |
| **Politics** | Travel | Travel | travel, travels, travelling, travelled, travelers, trip, trips, expedition, expeditions | 877 |
| **Politics** | Troops Withdrawal | Troops Withdrawal | withdraw troops, troops withdrawal | 47 |
| **Politics** | War | War | war, wars, combat, combating, airstrike, airstrikes, invasion, invasions, invade, invaded, invading, missile strikes, missile, missiles, rocket | 4,480 |
| **Science** | Discovery | Discovery | discover, discovers, discovered, discovering, discovery, discoveries | 594 |
| **Science** | Invention | Invention | invention, inventions, invent, invents, invented | 19 |
| **Social** | Ceremony | Ceremony | ceremony, ceremonies, celebrate, celebrates, celebrating, celebrated, festival, festivals, celebration, celebrations, commemorate, commemorates, procession, processions, funeral, funerals | 604 |
| **Social** | Coup | Coup | coup, coups, overthrow, overthrowed, overthrowing, overthrows | 331 |
| **Social** | Protest | Protest | protest, protests, protesting, protested, riot, riots, rioting, demonstrate, demonstrations, demonstrating, demonstrated, revolt, revolts, uprising, uprisings, rebellion, rebellions, strike, strikes, rally, rallies, protestors, demonstrators, rioters, violent clashes, mobs, mutiny, unrest, insurgency, insurgent, insurgents | 6,044 |
| **Social** | Revolution | Revolution | revolution, revolutions | 108 |